\documentclass{article}

\usepackage[dblblindworkshop, final]{neurips_2026}
\workshoptitle{Managing Agents that Manage Agents:
  Workshop on Responsible Use of Meta-Agents}
  
\usepackage[utf8]{inputenc} 
\usepackage[T1]{fontenc}    
\usepackage{hyperref}       
\usepackage{url}           
\usepackage{booktabs}       
\usepackage{amsfonts}      
\usepackage{nicefrac}    
\usepackage{microtype}    
\usepackage{xcolor}  
\usepackage{amsmath}
\usepackage{amssymb}
\usepackage{bbm}
\usepackage{enumitem}
\usepackage{multirow}
\usepackage{graphicx}
\usepackage{algorithm}
\usepackage{algorithmic}

\title{Habitats: A Framework for Open-Ended Evolution of Agent Objectives}

\author{%
Nicolas Bolouri \\
Google DeepMind \\
Montréal, Canada \\
\href{mailto:nb807@cam.ac.uk}
{\texttt{nb807@cam.ac.uk}}
\And
Gaya Mehenni \\
Google DeepMind \\
Montréal, Canada \\
\href{mailto:gmehenni@google.com}
{\texttt{gmehenni@google.com}}
\And
Daniel Toyama \\
Google DeepMind \\
Montréal, Canada \\
\href{mailto:kenjitoyama@google.com}
{\texttt{kenjitoyama@google.com}}
}
\begin{document}

\maketitle

\begin{abstract}
Designing what an agent should optimize can be as difficult as optimizing the agent itself. Existing language-model-based reward-generation methods typically refine objectives for predefined tasks and couple the decision of \emph{what} to explore with the synthesis of \emph{how} to implement it, potentially narrowing the resulting behavioral repertoire. We introduce \textbf{Habitats}, a framework for open-ended evolution of executable agent objectives. Habitats maintains a population of reward--policy pairs and evaluates every policy under every population reward, producing a cross-fitness matrix that exposes self-mastery, transfer, and functional diversity. A population-level Planner uses this information to select parent rewards and issue targeted mutation directives, while a separate Mutator translates those directives into validated reward programs. We evaluate Habitats in a discrete Card Sandbox and three continuous-control environments (Cart-Pole, Hopper, and Cheetah) against Independent Generation and Direct Mutation baselines. Across four environments and three random seeds, Habitats achieves the highest macro-averaged final Specialist Quality and Behavioral Diversity, with final diversity reaching $1.8\times$ that of Direct Mutation and $2.5\times$ that of Independent Generation. Finally, across ten downstream objectives unseen during evolution, frozen Habitats policy libraries achieve higher paired win rates than size-matched Direct Mutation and Independent Generation libraries at $10{,}000$ adaptation episodes, while also surpassing task-specific policies trained from scratch
for $20{,}000$ episodes. These results suggest that separating population-level planning from reward synthesis enables the discovery of diverse, learnable, and reusable behaviors.
\end{abstract}

%%%%%%%%%%%%%%%%%%%%%%%%%%%%%%%%%%%%%%%%%%%%%%%%%%%%%%%%%%%%%%%%
% Introduction
%%%%%%%%%%%%%%%%%%%%%%%%%%%%%%%%%%%%%%%%%%%%%%%%%%%%%%%%%%%%%%%%
\section{Introduction}

Reinforcement learning (RL) trains autonomous agents to maximize a scalar reward through interaction with an environment \citep{sutton2018reinforcement}. When that signal is well specified, reward maximization can induce complex and highly capable behavior, motivating the \emph{reward-is-enough} hypothesis that it may provide a sufficient basis for general intelligence \citep{silver2021reward}.

In open-ended domains, however, specifying the reward is itself the bottleneck. Common alternatives each fail differently: human preference labels are subjective and expensive at scale \citep{christiano2017deep,ouyang2022training}; learned reward models are brittle and vulnerable to overoptimization \citep{gao2023scaling}; and hand-designed rewards require domain-specific engineering while often admitting unintended shortcuts \citep{amodei2016concrete}. Progress in these settings may therefore depend as much on discovering useful objectives as on learning policies to optimize them. Furthermore, when future tasks are unknown, a diverse set of such objectives can produce a repertoire of behaviors that can later be reused \citep{barreto2019optionkeyboard}.

If objectives must themselves be discovered, evolutionary optimization provides a natural search mechanism because it does not require a differentiable objective. Recent work has used large language models (LLMs) as evolutionary operators over executable code, including mathematical constructions \citep{romeraparedes2024mathematical} and even reward functions \citep{ma2024eureka}. A common design, however, uses a single model both to select what to modify and to perform the modification, typically conditioning on the best candidates found so far. Selection pressure and the model's generative prior then reinforce one another, narrowing the population.

We introduce \textbf{Habitats}, a framework for open-ended objective evolution that separates population-level planning from reward synthesis. Each individual pairs an executable reward with its trained policy. An LLM-powered \emph{Planner} uses survival, cross-fitness, and diversity information to select parents and propose targeted mutation directives, which a separate LLM-based \emph{Mutator} converts into executable objectives. Offspring generation can therefore respond to population health rather than only refine the highest-scoring individuals. Across a discrete Card Sandbox and three continuous-control environments, we show that Habitats discovers diverse, high-quality, grounded behaviors that support sample-efficient adaptation to objectives unseen during evolution.

Our contributions are:

\begin{itemize}[leftmargin=10pt,topsep=1pt] 

\item A two-stage evolutionary architecture that separates population-level search decisions from reward synthesis, enabling diversity-aware evolution of executable objectives.

\item A behavior-grounded application of population-wide cross-fitness to objective evolution, measuring mastery, transfer, and diversity through learned policies rather than reward-code differences.

\item Evidence that population-guided planning improves behavioral diversity and grounded coverage while preserving specialist quality and enabling sample-efficient reuse under unseen objectives.

\end{itemize}

%%%%%%%%%%%%%%%%%%%%%%%%%%%%%%%%%%%%%%%%%%%%%%%%%%%%%%%%%%%%%%%%
% Related work
%%%%%%%%%%%%%%%%%%%%%%%%%%%%%%%%%%%%%%%%%%%%%%%%%%%%%%%%%%%%%%%%
\section{Related Work}

\paragraph{Reward design and learning from feedback.}
Reward shaping augments environment rewards with auxiliary signals that can accelerate learning while preserving the optimal policy, but still requires humans to specify both the desired behavior and shaping function \citep{ng1999policy}. Preference-based RL learns rewards from human comparisons, yet feedback is costly and learned reward models remain vulnerable to overoptimization \citep{christiano2017deep,ouyang2022training,gao2023scaling}. Habitats instead searches directly over executable objectives.

\paragraph{Coevolution, open-endedness, and quality diversity.}
Coevolutionary methods have long used population-wide interaction matrices and relative performance to preserve informative niches and sustain evolutionary progress \citep{rosin1997new,ficici1998challenges}; POET similarly pairs evolving environments with their optimized policies and transfers solutions between them \citep{wang2019poet}. Quality-diversity methods maintain collections of high-performing, behaviorally distinct solutions \citep{mouretclune2015illuminating,pughstanleysoros2016quality}. Separately, the Option Keyboard shows that policies learned from predefined pseudo-rewards can form a reusable basis for composing new behaviors \citep{barreto2019optionkeyboard}. Habitats builds on population-wide cross-evaluation but applies it to executable rewards, discovering objectives that induce a diverse, reusable policy repertoire rather than defining them in advance.

\paragraph{LLM-guided objective and agent design.}
Motif and MaestroMotif use LLM feedback to construct intrinsic rewards and skill libraries, while Eureka iteratively generates reward code using policy-training feedback \citep{klissarov2023motif,klissarov2024maestromotif,ma2024eureka}. FunSearch and AlphaEvolve evolve programs against predefined evaluators, and ADAS uses a meta-agent to construct agentic systems in code \citep{romeraparedes2024mathematical,lehman2023evolution,novikov2025alphaevolve,hu2025adas}. Habitats instead uses one prompted LLM role to monitor population-wide behavioral structure and another to synthesize its proposed executable objectives.

%%%%%%%%%%%%%%%%%%%%%%%%%%%%%%%%%%%%%%%%%%%%%%%%%%%%%%%%%%%%%%%%
% Methods
%%%%%%%%%%%%%%%%%%%%%%%%%%%%%%%%%%%%%%%%%%%%%%%%%%%%%%%%%%%%%%%%
\section{The Habitats Framework}
\label{sec:methods}

Habitats casts open-ended objective discovery as an evolutionary search over a population of executable reward functions. Rather than optimizing a single hand-specified reward, Habitats maintains an evolving population of individuals, each pairing a synthesized reward with a policy trained against it. We first outline the system lifecycle (\S\ref{subsec:lifecycle}), then formalize its core objects (\S\ref{subsec:definitions}), define cross-fitness evaluation, survivor selection, and quality-diversity tracking (\S\ref{subsec:cross_fitness}), describe Planner--Mutator offspring synthesis (\S\ref{subsec:dual_llm}), and conclude with implementation details (\S\ref{subsec:implementation}).

% New subsection
\subsection{System Overview and Lifecycle}
\label{subsec:lifecycle}

\begin{figure}[t]
    \centering
    \includegraphics[width=\linewidth]{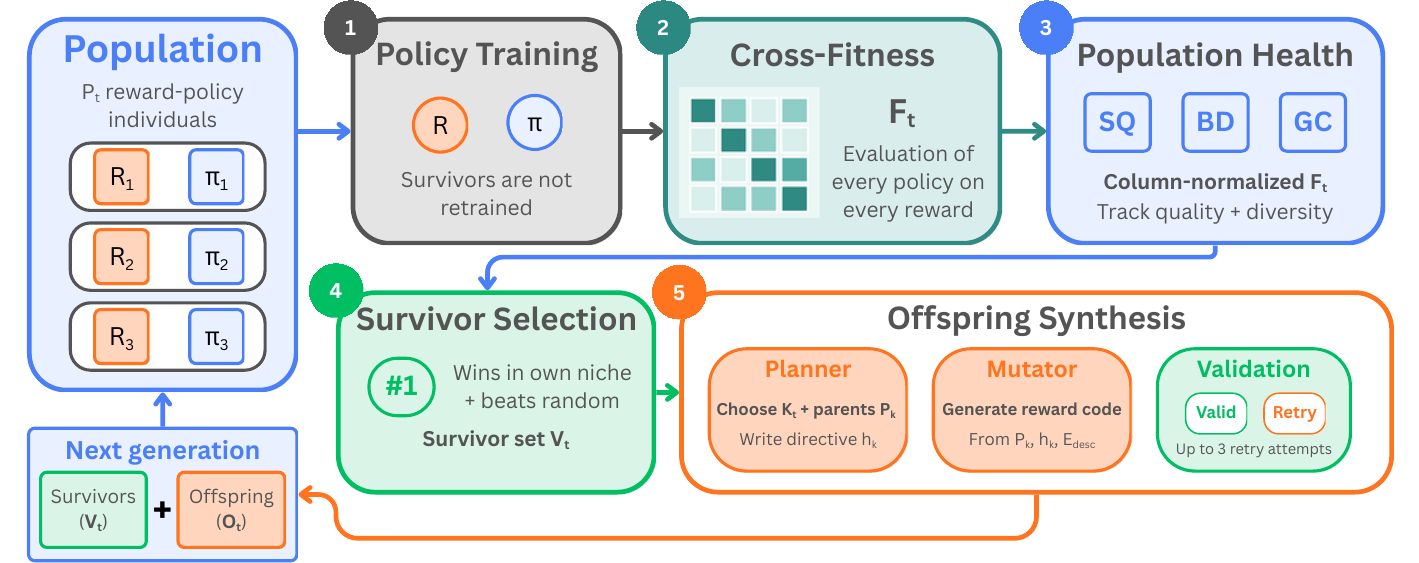}
    \vspace{-6pt}
    \caption{
        \textbf{Habitats evolutionary lifecycle.} At generation $t$, the population $\mathcal{P}_t$ contains executable reward--policy pairs. \textbf{(1)} Policies for newly synthesized rewards are trained from scratch, while inherited survivor policies remain fixed. \textbf{(2)} Every policy is evaluated under every population reward to construct the cross-fitness matrix $\mathbf{F}_t$. \textbf{(3)} Normalized cross-fitness profiles are used to compute Specialist Quality (SQ), Behavioral Diversity (BD), and Grounded Behavioral Profile Coverage (GC). \textbf{(4)} Individuals whose policies dominate their own reward niches and outperform the random-policy baseline form the survivor set $\mathcal{V}_t$. \textbf{(5)} The Planner determines the offspring count, selects parents, and proposes mutation directives, while the Mutator generates and validates executable reward programs. Valid offspring join the survivors to form $\mathcal{P}_{t+1}=\mathcal{V}_t\cup\mathcal{O}_t$; invalid candidates receive up to three replacement attempts before their offspring slot is discarded.
    }
    \label{fig:fig1}
\end{figure}

An evolutionary run proceeds through generations $t\in\{1,\dots,T\}$. At each generation $t<T$, Habitats follows the five-stage lifecycle illustrated in Fig.~\ref{fig:fig1}. Newly synthesized rewards are paired with policies trained from scratch, while surviving policies retain their parameters. Every policy is evaluated under every population reward to construct a cross-fitness matrix and population-level quality-diversity statistics. An individual survives only if its policy dominates its own reward and outperforms the random-policy baseline. The Planner then selects parents and issues mutation directives (natural-language specifications of intended behavioral changes) which the Mutator implements as validated reward programs. Survivors and valid offspring form $\mathcal{P}_{t+1}$. Algorithm~\ref{alg:habitats} provides the complete procedure.

% New subsection
\subsection{Problem Formalization and Key Definitions}
\label{subsec:definitions}

Evolution proceeds in generations, each comprising one full pass through the five-step lifecycle above. At generation $t$, Habitats maintains a population $\mathcal{P}_t=\{I_1,\dots,I_{N_t}\}$ of $N_t$ individuals. Each individual pairs an executable reward function with the policy trained against it: $ I_i = (R_i,\pi_i)$. Here, $R_i$ defines the objective and $\pi_i(a\mid s)$ is the corresponding policy. Surviving reward--policy pairs are carried forward across generations with their learned policy parameters fixed; only newly synthesized rewards are paired with newly initialized policies and trained using the environment's fixed policy-optimization procedure. In other words, only reward programs are subject to evolutionary variation.

\paragraph{Reward function contract.}
Each objective is represented as an executable Python function
\begin{equation}
R_i : (s,a,s',\text{info}) \longrightarrow \mathbb{R},
\label{eq:reward_contract}
\end{equation}
where $s,s'\in\mathcal{S}$ are the pre- and post-transition states, $a\in\mathcal{A}$ is the selected action, and $\text{info}$ contains auxiliary environment metadata (see Appendix~\ref{app:environment_specs} for the environment-specific reward interfaces and exposed metadata). The environment dynamics $\mathcal{T}(s'\mid s,a)$ remain fixed throughout evolution; only $R_i$ evolves, and it governs the training of $\pi_i$. Habitats is agnostic to the specific control environment: a new environment can be integrated through a lightweight adapter exposing its state, action, transition, and metadata interfaces without modifying the evolutionary loop.

Although written in transition-level notation, reward programs may maintain
internal state across steps, enabling history-dependent objectives; this state is reset at episode boundaries and isolated across evaluation rollouts.

% Fig 1

% New subsection
\subsection{Cross-Fitness Evaluation and Survivor Selection}
\label{subsec:cross_fitness}

\paragraph{Cross-fitness evaluation.}
Habitats evaluates every policy against every reward in the population, producing an $N_t\times N_t$ cross-fitness matrix $\mathbf{F}_t\in\mathbb{R}^{N_t\times N_t}$:
\begin{equation}
\mathbf{F}_t[i,j]
=
\operatorname{Eval}(\pi_i,R_j),
\label{eq:crossfitness}
\end{equation}
where $\mathbf{F}_t[i,j]$ is the performance of policy $\pi_i$ at generation $t$, trained against reward $R_i$, when evaluated under reward $R_j$. The diagonal $\mathbf{F}_t[i,i]$ measures \emph{self-mastery}: how well a policy satisfies the objective against which it was trained. The off-diagonal entries $\mathbf{F}_t[i,j]$, $i\neq j$, measure \emph{transferability}: how well that behavior generalizes to objectives for which it was not explicitly trained.

Because synthesized rewards may have incomparable numerical scales, each column of $\mathbf{F}_t$ is rescaled to $[0,1]$, mapping its population minimum and maximum to $0$ and $1$. We denote the resulting matrix by $\tilde{\mathbf{F}}_t$. This prevents reward magnitude from substituting for reward quality.

The complete cross-fitness evaluation procedure, including the cross-fitness matrix normalization formula, rollout counts, episode horizons, aggregation, and random-policy evaluation, is provided in Appendix~\ref{app:evaluation}.

\paragraph{Survivor selection.}
Survivors are selected through niche dominance. Let $\mathbf{F}_t[\mathrm{rand},i]$ denote the fitness of a random policy under reward $R_i$. An individual $I_i$ survives if and only if its policy achieves the highest fitness on its own reward niche relative to all co-evolving policies and outperforms the random-policy baseline:
\begin{equation}
\operatorname{Survivor}_t(I_i)
=
\mathbf{1}
\left(
\mathbf{F}_t[i,i]
=
\max_k \mathbf{F}_t[k,i]
\quad\text{and}\quad
\mathbf{F}_t[i,i]
>
\mathbf{F}_t[\mathrm{rand},i]
\right).
\label{eq:selection}
\end{equation}
Because policy training is finite-budget, diagonal dominance is not automatic: a policy trained on another reward may transfer better to $R_i$. The criterion therefore retains rewards that are learnable within the fixed budget and induce their strongest population policy.
The surviving individuals form the set $\mathcal{V}_t\subseteq\mathcal{P}_t$.

\paragraph{Quality-diversity tracking.}
To monitor population health and detect convergence toward homogeneous behavior, Habitats tracks two complementary quality-diversity metrics computed from the normalized cross-fitness matrix $\tilde{\mathbf{F}}_t$.

\textbf{Specialist Quality (SQ)} measures the average proficiency of policies on their own synthesized reward niches:
\begin{equation}
    \mathrm{SQ}_t
    =
    \frac{1}{|\mathcal{P}_t|}
    \sum_{i=1}^{N_t}
    \tilde{\mathbf{F}}_t[i,i].
    \label{eq:sq}
\end{equation}
\textbf{Behavioral Diversity (BD)} measures the functional distinctness of the
policies in the population. Let $\mathbf{f}_t[i] = \tilde{\mathbf{F}}_t[i,:]$ denote the normalized cross-fitness profile of policy $\pi_i$ at generation $t$. Before computing BD, zero-norm profiles are excluded and the remaining profiles are reindexed; let $N_t^{\mathrm{BD}}$ denote their number. Note that no
zero-norm profiles occurred in the reported experiments. Behavioral Diversity
is the average pairwise cosine distance between the remaining profiles:
\begin{equation}
    \mathrm{BD}_t
    =
    \frac{2}{N_t^{\mathrm{BD}}(N_t^{\mathrm{BD}}-1)}
    \sum_{1 \leq i < j \leq N_t^{\mathrm{BD}}}
    \left(
        1 -
        \frac{
            \mathbf{f}_t[i] \cdot \mathbf{f}_t[j]
        }{
            \|\mathbf{f}_t[i]\|_2
            \|\mathbf{f}_t[j]\|_2
            + \epsilon
        }
    \right).
    \label{eq:bd}
\end{equation}
If $N_t^{\mathrm{BD}}<2$, we define $\mathrm{BD}_t=0$. Here,
$\epsilon>0$ provides numerical stability. High $\mathrm{BD}_t$ indicates that
the population contains policies with distinct functional specializations
across the evolved reward landscape.

% New subsection
\subsection{Offspring Synthesis: Planner and Mutator}
\label{subsec:dual_llm}

Naive genetic-programming operators, such as random token substitution or subtree crossover, frequently produce invalid syntax or incoherent reward programs. They also provide no mechanism for steering mutation toward underexplored regions of the population's behavior space. Habitats addresses both limitations with a two-stage architecture: a \emph{Planner} decides what to explore based on population-level state, while a separately prompted \emph{Mutator} implements that decision as reward code.

\paragraph{The Planner ($\mathcal{M}_{\mathrm{plan}}$).}
The Planner inspects the current survivors $\mathcal{V}_t$, the normalized cross-fitness matrix $\tilde{\mathbf{F}}_t$, and the population quality-diversity metrics $(\mathrm{SQ}_t,\mathrm{BD}_t)$. It determines the number of offspring $K_t$ and, for each offspring slot $k\in\{1,\ldots,K_t\}$, selects a parent set $P_k\subseteq\mathcal{V}_t$, with $0\leq|P_k|\leq 3$, and produces a natural-language mutation directive $h_k$:
\begin{equation}
(P_k,h_k)
=
\mathcal{M}_{\mathrm{plan}}
\left(
\mathcal{V}_t,
\tilde{\mathbf{F}}_t,
\mathrm{SQ}_t,
\mathrm{BD}_t
\right).
\label{eq:planner}
\end{equation}
This centralized planning step is the mechanism through which population-wide information, including survivor fitness profiles, specialist quality, and behavioral diversity, directly guides parent selection and offspring generation.

\paragraph{The Mutator ($\mathcal{M}_{\mathrm{mutate}}$).}
For each offspring slot $k$, the Mutator receives the environment interface description $E_{\mathrm{desc}}$, the reward functions associated with the selected parent set $P_k$, and the Planner directive $h_k$. It synthesizes a candidate reward function
\begin{equation}
R_{\mathrm{child}}^{(k)}
=
\mathcal{M}_{\mathrm{mutate}}
\left(
E_{\mathrm{desc}},
\{R_p : I_p \in P_k\},
h_k
\right).
\label{eq:mutator}
\end{equation}
Each valid child reward is paired with a newly initialized policy trained against it, forming an offspring individual
$I_{\mathrm{child}}^{(k)}=(R_{\mathrm{child}}^{(k)},\pi_{\mathrm{child}}^{(k)})$.
We denote the set of valid individuals produced across the $K_t$ offspring
slots by $\mathcal{O}_t$. The next population is then
$\mathcal{P}_{t+1}=\mathcal{V}_t\cup\mathcal{O}_t$.

Every synthesized reward program undergoes static syntax checking and sandboxed test execution. If validation fails, the candidate is rejected and the Mutator is called again, for up to three replacement attempts. If all four attempts fail, the offspring slot is discarded and no individual is added to
$\mathcal{O}_t$.

% New subsection
\subsection{Implementation and Hyperparameters}
\label{subsec:implementation}

The complete set of prompts used by the system during initialization, planning, and mutation is included in Appendix~\ref{app:prompts}.
Complete policy-optimization, evaluation, language-model, software, compute,
and random-seed details are provided in Appendix~\ref{app:hyperparameters}.

%%%%%%%%%%%%%%%%%%%%%%%%%%%%%%%%%%%%%%%%%%%%%%%%%%%%%%%%%%%%%%%%
% Experimental Setup
%%%%%%%%%%%%%%%%%%%%%%%%%%%%%%%%%%%%%%%%%%%%%%%%%%%%%%%%%%%%%%%%
\section{Experimental Setup}
\label{sec:experiments}

We evaluate Habitats in four control environments: a discrete Card Sandbox and three continuous-control domains from the publicly available DeepMind Control Suite \citep{tassa2018deepmind}, which uses the MuJoCo physics engine. In order to parallelize training efficiently, we also utilize the Brax engine \citep{freeman2021braxdifferentiablephysics}. The Card Sandbox uses tabular Q-learning, while the MuJoCo environments use Proximal Policy Optimization (PPO) \citep{schulman2017proximal}. In all environments, policies are trained exclusively against synthesized rewards; native task rewards are not used for policy optimization. The full environment specifications are provided in Appendix~\ref{app:environment_specs}.

% New subsection
\subsection{Environments}
\label{subsec:setup_environments}

\paragraph{Card Sandbox.}
The Card Sandbox is a discrete card-management environment in which agents manipulate cards across a deck, hand, and table using nine primitive actions: pass, draw, play to one of three table slots,
retrieve from one of three table slots, and discard. The environment provides no predefined task objective, so all learned behavior must arise from synthesized rewards. Each run begins with $N_\text{init}=3$ reward--policy pairs and proceeds for $T=15$ generations, which we find sufficient in practice (see Section~\ref{subsec:robustness_analysis}). Policies are represented by tabular Q-functions.

\paragraph{MuJoCo continuous control.}
We evaluate continuous-control behavior in Cart-Pole, Hopper, and Cheetah. These environments span balancing and locomotion tasks with continuous state and action spaces. Each run begins with $N_\text{init}=4$ reward--policy pairs and proceeds for $T=6$ generations, which we find sufficient in practice
(see Section~\ref{subsec:robustness_analysis}). All policies are trained using PPO with a fixed architecture, optimizer, and interaction budget.

% New subsection
\subsection{Evaluation Protocol}
\label{subsec:setup_protocol}

We perform three independent evolutionary runs per environment and method using random seeds. All methods use the same initial population sizes, generation counts, and policy-level hyperparameters. Every newly generated policy receives the
same environment-interaction budget, and every policy--reward pair is evaluated
the same way. Because population sizes evolve endogenously, total policy-training interactions and cross-fitness evaluations may differ across methods.

% New subsection
\subsection{Baselines}
\label{subsec:setup_baselines}

We compare Habitats against two competitive evolutionary baselines with matched per-policy training and evaluation budgets (see Appendix~\ref{app:baselines} for full details):

\begin{itemize}[leftmargin=*,noitemsep,topsep=2pt]
\item \textbf{Independent Generation.}
At each generation, offspring rewards are generated independently without parent conditioning or Planner guidance, while retaining the same evolutionary selection and elitism.

\item \textbf{Direct Mutation.}
This is Habitats without the Planner: surviving rewards are passed directly to the Mutator, preserving lineage-based inheritance while removing population-level guidance.
\end{itemize}

% New subsection
\subsection{Metrics}
\label{subsec:setup_metrics}

\paragraph{Specialist Quality (SQ).}
Specialist Quality is the mean normalized diagonal of the cross-fitness matrix, as defined in Eq.~\eqref{eq:sq}. We report its mean across generations and its value on $\mathcal{P}_T$, along with its equal-weight macro-averages across environments.

\paragraph{Behavioral Diversity (BD).}
Behavioral Diversity is the mean pairwise cosine distance between normalized cross-fitness profiles, as defined in Eq.~\eqref{eq:bd}. We report its mean across generations and its value on $\mathcal{P}_T$, along with its equal-weight macro-averages across environments.

\paragraph{Grounded Behavioral Profile Coverage (GC).}
GC counts grounded, functionally distinct policies within each generation. A
policy is grounded when its reward executes successfully and its self-fitness
exceeds the corresponding random-policy baseline by
\[
L_i
=
\frac{\mathbf{F}_t[i,i]-\mu_i^{\mathrm{rand}}}
{\sigma_i^{\mathrm{rand}}+\epsilon}
>
\lambda,
\qquad \lambda=0.5.
\]
Among grounded policies, normalized cross-fitness profiles are rounded to one
decimal place and deduplicated within each generation; their number defines
$\mathrm{GC}_t$. We report final-generation coverage and
$\mathrm{GC}_{\mathrm{AUC}}=\sum_{t=1}^{T}\mathrm{GC}_t$.

Full metric definitions and edge-case handling are provided in Appendix~\ref{app:evaluation}.

% New subsection
\subsection{Downstream Adaptation}
\label{subsec:downstream_adaptation}

To test whether evolved behaviors remain useful beyond their original objectives, we evaluate frozen policy libraries from Habitats, Direct Mutation, and Independent Generation on ten Card Sandbox tasks fixed before evolution. For each method and seed, we uniformly sample 15 final survivors without reference to downstream performance. A meta-policy routes among these frozen policies and is trained for $1{,}000$, $2{,}000$, $5{,}000$, $10{,}000$, or $20{,}000$ episodes, then compared with a task-specific Scratch policy trained for $20{,}000$ episodes using matched hyperparameters and seeds. We aggregate performance using paired win rate. Full task definitions, evaluation details, and per-task results are provided in Appendix~\ref{app:meta_evaluation}.

% Fig. 2
\begin{figure}[t]
    \centering
    \includegraphics[width=\linewidth]{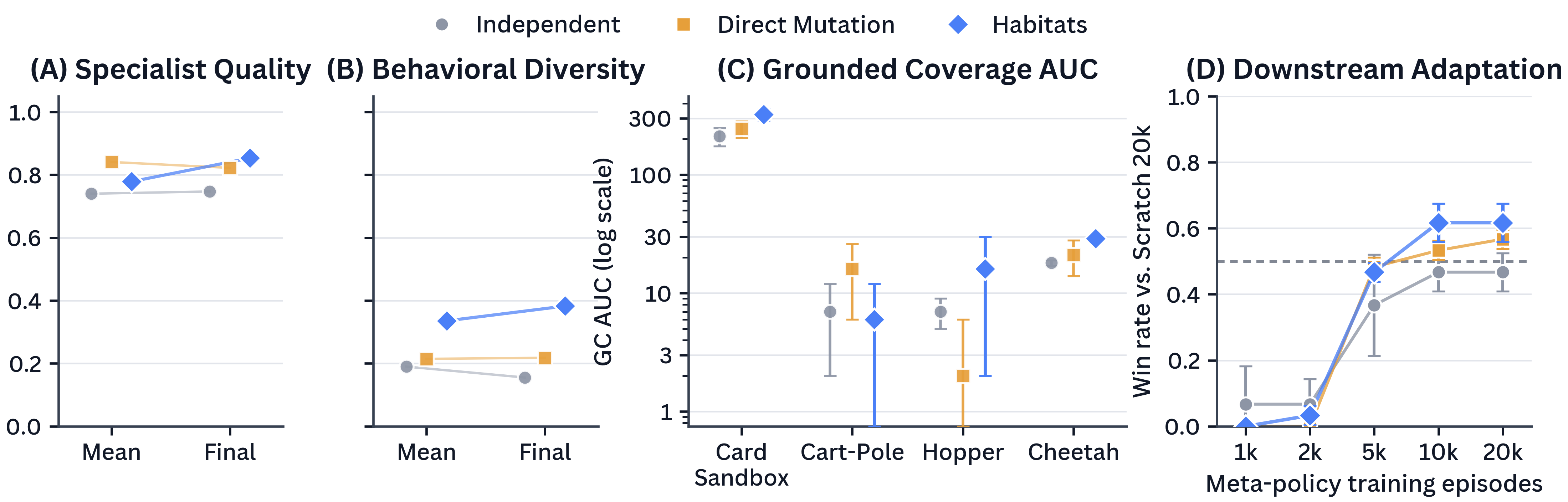}
    \vspace{-10pt}
    \caption{
    \textbf{Population quality, diversity, grounded coverage, and downstream adaptation.}
    \textbf{(A--B)} SQ and BD are equal-weight macro-averages across environments,
    shown as their means across generations and their values in the final population.
    \textbf{(C)} $\mathrm{GC}_{\mathrm{AUC}}$ is reported separately for each
    environment on a logarithmic scale. \textbf{(D)} Downstream adaptation is
    measured by paired win rate against task-specific Scratch policies trained for
    $20{,}000$ episodes. Meta-policies route among size-matched, frozen libraries
    produced by Habitats, Direct Mutation, or Independent Generation and are trained
    for $1{,}000$, $2{,}000$, $5{,}000$, $10{,}000$, or $20{,}000$ episodes; the
    dashed line denotes parity. Error bars in \textbf{(C--D)} indicate
    $\pm 1$ standard deviation over three seeds.
    }
    \label{fig:fig2}
\end{figure}

%%%%%%%%%%%%%%%%%%%%%%%%%%%%%%%%%%%%%%%%%%%%%%%%%%%%%%%%%%%%%%%%
% Results
%%%%%%%%%%%%%%%%%%%%%%%%%%%%%%%%%%%%%%%%%%%%%%%%%%%%%%%%%%%%%%%%
\section{Results and Analysis}
\label{sec:results}

\subsection{Overall Objective-Discovery Performance}
\label{subsec:overall_results}

Figure~\ref{fig:fig2}A-C summarizes the principal results, with complete per-environment values reported in Appendix~\ref{app:full_results} (Tables~\ref{tab:full_sq_bd_results} and~\ref{tab:full_gc_results}). The baselines are deliberately stringent ablations of Habitats: all three methods share the same evolutionary loop, policy training, cross-fitness evaluation, and survivor selection, but differ in how offspring rewards are generated and allocated. Independent Generation removes parent conditioning and population-level planning, whereas Direct Mutation retains parent-conditioned evolution but removes the Planner. Both baselines allocate one offspring slot per survivor, whereas Habitats allows the Planner to adaptively determine the offspring count from the current population state.

\paragraph{Specialist Quality.}
All three methods produce policies that specialize effectively under their paired rewards. Direct Mutation achieves the highest macro-averaged mean SQ across generations ($0.8409$), followed by Habitats ($0.7787$) and Independent Generation ($0.7400$). On the final evaluated population, however, Habitats achieves the highest macro-average ($0.8533$), compared with $0.8222$ for Direct Mutation and $0.7471$ for Independent Generation. Both Direct Mutation and Habitats exceed Independent Generation in mean SQ in all four environments and in final SQ in three of four, consistent with evolutionary inheritance helping preserve and refine productive objectives. The per-environment results are mixed: Direct Mutation leads mean SQ in Card Sandbox, Cart-Pole, and Hopper, while Habitats leads in Cheetah; the best final SQ is distributed across all three methods. Habitats therefore does not uniformly dominate specialist quality, but remains competitive while optimizing for a broader population-level objective.

\paragraph{Behavioral Diversity.}
The clearest advantage of Habitats is behavioral diversity. Its macro-averaged mean BD is $0.3355$, exceeding Direct Mutation ($0.2146$) by $56\%$ and Independent Generation ($0.1899$) by $77\%$. The difference grows on the final population: Habitats reaches $0.3825$, approximately $1.8\times$ Direct Mutation ($0.2180$) and $2.5\times$ Independent Generation ($0.1552$). Habitats achieves the highest final BD in all four environments and the highest mean BD in three of four, trailing Direct Mutation only slightly in Cart-Pole ($0.2987$ versus $0.3163$). The largest final-population differences appear in Hopper and Cheetah, where Habitats produces substantially more distinct cross-fitness profiles. The effect is most pronounced in Hopper, where Habitats achieves $7.6\times$ the mean BD and $4.8\times$ the final BD of Direct Mutation. Because Direct Mutation differs from Habitats primarily by removing the Planner and its population-level context, this pattern is consistent with diversity-aware planning steering mutation toward less redundant behavioral niches.

\paragraph{Grounded coverage.}
The coverage results show that these diversity gains generally correspond to distinct, learnable behaviors rather than merely different reward programs. Habitats achieves the highest $\mathrm{GC}_{\mathrm{AUC}}$ in Card Sandbox, Hopper, and Cheetah, while Direct Mutation leads in Cart-Pole. On the final population, Habitats attains the highest mean GC in Card Sandbox, Hopper, and Cheetah and ties Direct Mutation in Cart-Pole. Coverage varies substantially across seeds and cannot be meaningfully macro-averaged because its scale depends on the evolutionary horizon and population size. Nevertheless, its per-environment pattern reinforces the central result: \textbf{population-guided planning primarily improves behavioral diversity and grounded coverage while preserving competitive specialist quality.
}

% Subsection 2
\subsection{Downstream Utility of Evolved Policy Libraries}
\label{subsec:results_downstream_adaptation}

Figure~\ref{fig:fig2}D shows that Habitats supports sample-efficient adaptation to unforeseen objectives. Across ten tasks held out from evolution, a meta-policy routing among 15 frozen Habitats policies achieves a paired win rate of $0.467 \pm 0.029$ after $5{,}000$ episodes, approaching mean parity with task-specific Scratch policies trained for $20{,}000$ using only one quarter of the downstream-training budget. At $10{,}000$ episodes, it reaches $0.617 \pm 0.058$, surpassing parity using half the training budget. Under the same protocol with size-matched libraries, this exceeds Direct Mutation ($0.533 \pm 0.029$) and Independent Generation ($0.467 \pm 0.058$), and remains higher than either baseline after $20{,}000$ episodes.

These results show that Habitats discovers non-trivial skills that remain useful under objectives unknown during evolution. Its advantage over size-matched Direct Mutation and Independent Generation libraries suggests that population-level planning produces a more broadly reusable repertoire, rather than merely a larger collection of task-specific behaviors. Because the library is evolved once and reused, its discovery cost can be amortized across downstream tasks, enabling new meta-policies to adapt efficiently by learning how to route among existing skills. Complete protocols and results are provided in Appendix~\ref{app:meta_evaluation}.

% Fig. 3
\begin{figure}[t]
    \centering
    \includegraphics[width=\linewidth]{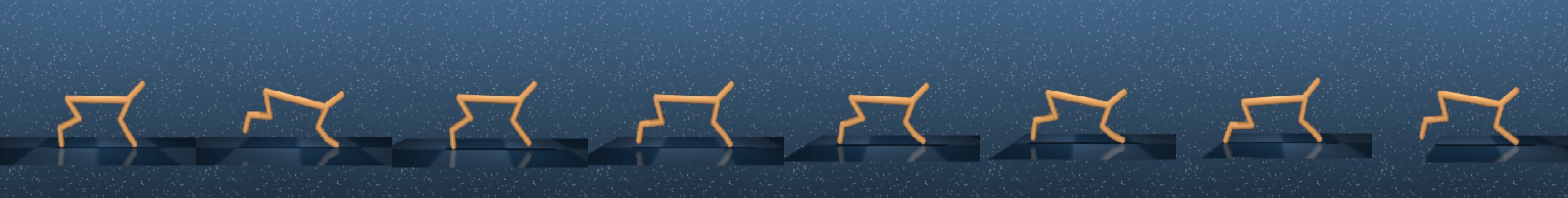}
    \vspace{-10pt}
    \caption{
       \textbf{Evolved backward locomotion.} In the Cheetah environment, Habitats produced an individual that learned to move backward in response to the Planner directive \emph{Smooth Retro Runner: High-Speed Coordinated Backward Gallop}. \href{https://www.youtube.com/shorts/y8KB_aivMMw}{Link} to video.
    }
    \vspace{-10pt}
    \label{fig:fig3}
\end{figure}

% Subsection 3
\subsection{Qualitative Analysis of Evolved and Discovered Behaviors}
\label{subsec:results_qualitative}

The quantitative results show that Habitats maintains distinct behavioral niches; we next examine what those niches contain. Across discrete and continuous-control environments, evolution produced interpretable objectives that exploit different affordances of the same embodiment without receiving a target behavior or predefined skill taxonomy.

\paragraph{Distinct uses of the same embodiment.}
Figure~\ref{fig:fig3} shows the clearest example. In response to the Planner directive \emph{Smooth Retro Runner: High-Speed Coordinated Backward Gallop}, the Mutator constructed an executable objective whose policy learned stable, coordinated backward locomotion. Other Cheetah objectives produced a low forward crawl or repeated vertical jumps with little horizontal displacement. Hopper policies similarly ranged from explosive forward leaps to long, controlled strides that achieved speed without rapid joint motion. These behaviors differ in direction, posture, and movement strategy rather than representing minor variations of a conventional gait.

\paragraph{Compositional objective discovery.}
The Card Sandbox exhibits a parallel pattern in a discrete domain. Evolution progressed from simple relations, such as the ascending arrangement required by \emph{The Sorting Table}, to increasingly structured mini-games. Later objectives added equal spacing, fixed that spacing to one, or combined spatial patterns with primality and divisibility constraints. Habitats therefore composed ordering, position, parity, and arithmetic relations rather than repeatedly expressing the same underlying task.

\paragraph{Contraction and recovery.}
One Card Sandbox run exposes the population-level feedback loop particularly clearly. A policy induced by \emph{The Composite Arithmetic Solver} generalized across several existing objectives and displaced multiple specialists, temporarily contracting the population around a narrower set of behaviors. Subsequent Planner--Mutator cycles generated new objectives combining sandwich patterns, primality, parity, and divisibility, restoring a broader repertoire of specializations. The trace therefore captures a complete qualitative cycle: specialization, generalist takeover, population contraction, and renewed diversification. This recovery illustrates the intended role of population-level planning: offspring generation responds to the state of the evolving ecology rather than repeatedly mutating its dominant individual in isolation.

Together, these examples show that Habitats discovers meaningful skills and structural concepts while responding dynamically to population contraction. Selection operates on what each reward causes a policy to learn, and the Planner uses that behavioral context to redirect subsequent objective generation. Appendix~\ref{app:qualitative_examples} provides many additional Cheetah and Hopper behaviors, Card Sandbox objective families, evolutionary progressions, and cross-niche replacement events.

% Subsection 4
\subsection{Robustness to Evolutionary Horizon and Language-Model Backbone}
\label{subsec:robustness_analysis}

\paragraph{Evolutionary horizon.}
To test whether our results depend on the chosen evolutionary horizon, we
extend Card Sandbox from $15$ to $30$ generations and all three
continuous-control environments from $6$ to $10$. By the original stopping
points, SQ is already near its long-run level, and Habitats continues to
achieve the highest BD throughout the additional generations in all four
environments. Extending evolution therefore does not change the main comparison
between methods, supporting the original horizons as sufficient for evaluation.
Complete trajectories are provided in Appendix~\ref{app:horizon_analysis}.

\paragraph{Language-model backbone.}
To assess the effect of the underlying language model, we evaluate Habitats in Card Sandbox using \texttt{gemini-3.5-flash},
\texttt{gemini-3.8-flash}, and \texttt{gemini-4-argon}, while holding all
other settings fixed. All three achieve near-perfect final SQ, showing that
Habitats reliably discovers learnable objectives across backbones. The
strongest tested backbone, \texttt{gemini-4-argon}, produces substantially
greater behavioral diversity, reaching a final BD of $0.7959 \pm 0.0577$:
$62\%$ higher than \texttt{gemini-3.5-flash} and $110\%$ higher than
\texttt{gemini-3.8-flash}. Crucially, this increase occurs without reducing
specialist quality. This supports the central design of Habitats: stronger
Planner and Mutator capabilities can be translated into broader populations
of distinct, learnable behaviors, suggesting further headroom as language
models improve. Complete results and evolutionary trajectories are provided
in Appendix~\ref{app:model_ablation}.

%%%%%%%%%%%%%%%%%%%%%%%%%%%%%%%%%%%%%%%%%%%%%%%%%%%%%%%%%%%%%%%%
% Discussion
%%%%%%%%%%%%%%%%%%%%%%%%%%%%%%%%%%%%%%%%%%%%%%%%%%%%%%%%%%%%%%%%
\section{Discussion}

\paragraph{Population-level planning changes the character of evolution.}
Habitats does not uniformly maximize specialist quality: Direct Mutation achieves the highest mean SQ across generations, whereas Habitats achieves the highest final macro-average. Direct Mutation's advantage over Independent Generation suggests that parent conditioning provides useful evolutionary memory. Habitats' principal advantage lies in population structure: it achieves the highest final BD in all four environments and the highest grounded coverage in three of four while maintaining competitive SQ. This pattern suggests that parent-conditioned mutation refines productive objectives, while population-level planning prevents that refinement from collapsing onto a narrow behavioral family. The largest diversity gains occur in Hopper and Cheetah, whose embodiments support multiple gaits, postures, and movement directions. Yet Cart-Pole also achieves its highest final SQ and BD under Habitats, indicating that population guidance can benefit constrained environments. Determining when these gains justify the additional planning overhead remains a direction for future work.

\paragraph{Cross-fitness grounds novelty in learned behavior.}
Building on coevolutionary cross-evaluation, Habitats uses cross-fitness to distinguish reward novelty from behavioral novelty: syntactically different rewards need not induce distinct policies. This distinction is illustrated by the observed generalist takeover: a broadly capable policy displaced several specialists and temporarily contracted the population, after which population-level planning helped restore behavioral breadth. Habitats is therefore better understood as an ecology of competing objectives and behaviors than as a collection of independently generated rewards. The resulting crawling, jumping, bounding, and striding behaviors further suggest an emergent objective basis analogous to the predefined cumulants of the Option Keyboard \citep{barreto2019optionkeyboard}.

\paragraph{When is objective evolution preferable?}
Task-specific learning remains appropriate when a single objective is known in advance. Population-guided evolution is more relevant when desired capabilities are unknown or plural and the goal is to construct a repertoire that can be inspected, composed, or reused. The Card Sandbox demonstrates this utility: routing among frozen Habitats policies approaches mean parity with task-specific policies trained from scratch for $20{,}000$ episodes after only $5{,}000$ episodes and surpasses it after $10{,}000$. Moreover, the Habitats library adapted for $10{,}000$ episodes achieves a higher mean win rate than size-matched Direct Mutation and Independent Generation libraries adapted for $20{,}000$, indicating that population-level planning produces a more reusable behavioral repertoire. Because the library is evolved once and reused, its discovery cost can be amortized across unforeseen objectives. By evolving the objectives that shape behavior, Habitats moves toward open-ended objective discovery; its bounded environments, interfaces, and evolutionary horizons, however, preclude claims of unbounded open-ended evolution.

\paragraph{Limitations and outlook.}\label{limitations}
Our evaluation covers four simulated environments, three evolutionary seeds,
and finite horizons; the language-model ablation is limited to Card Sandbox
and three Gemini backbones. Although per-policy budgets are matched, total
computation varies with endogenous population size. Because each method
generates its own rewards, BD and GC use method-specific cross-fitness spaces
whose coordinates also change across generations; they therefore characterize
within-population structure rather than behavior under a shared reward basis.
The size-matched downstream evaluation provides a common external comparison
of library utility. Finite-rollout survivor selection may also discard near
ties. Habitats evolves only rewards while holding the policy architecture and
optimization procedure fixed, even though different objectives may benefit
from different policies. Future work could jointly evolve objectives and
policies and extend survivor selection to consider computational efficiency,
allowing a comparably capable but cheaper policy to displace a more expensive
specialist. Moreover, providing the Planner with explicit population context
may become impractical as populations and histories grow. Further directions
include compressed population representations, persistent behavioral archives,
broader environments, compute-amortized downstream evaluation, and the joint
evolution of objectives, environments, and compositional controllers.

Within these limits, our results show that coupling language models with behavioral evaluation and population-level planning can produce evolving objective repertoires whose policies remain diverse, grounded, and reusable.

%%%%%%%%%%%%%%%%%%%%%%%%%%%%%%%%%%%%%%%%%%%%%%%%%%%%%%%%%%%%%%%%
% Responsible-Use Statement and Acknowledgments
%%%%%%%%%%%%%%%%%%%%%%%%%%%%%%%%%%%%%%%%%%%%%%%%%%%%%%%%%%%%%%%%
\section*{Responsible-Use Statement}

Habitats automates the generation of executable agent objectives; outside
controlled simulations, this capability could produce unintended goals,
exploit evaluation flaws, or reduce human oversight over agent behavior. Our
experiments are confined to sandboxed simulations with retained code and
execution traces. Real-world use would require sandboxing, restricted
permissions, independent evaluation, human approval before deployment, and
reliable halt and rollback mechanisms.

\section*{Acknowledgments}

We thank Yousuf Ali for his assistance in open-sourcing the Habitats codebase,
available at \url{https://github.com/google-deepmind/habitats}.

%%%%%%%%%%%%%%%%%%%%%%%%%%%%%%%%%%%%%%%%%%%%%%%%%%%%%%%%%%%%%%%%
% References & Transition
%%%%%%%%%%%%%%%%%%%%%%%%%%%%%%%%%%%%%%%%%%%%%%%%%%%%%%%%%%%%%%%%
\newpage

\bibliographystyle{plainnat}
\bibliography{references}

\newpage
\appendix

%%%%%%%%%%%%%%%%%%%%%%%%%%%%%%%%%%%%%%%%%%%%%%%%%%%%%%%%%%%%%%%%
% New Appendix
%%%%%%%%%%%%%%%%%%%%%%%%%%%%%%%%%%%%%%%%%%%%%%%%%%%%%%%%%%%%%%%%
\section{Habitats Algorithm}
\label{app:algorithm}

Algorithm~\ref{alg:habitats} formalizes the evolutionary lifecycle introduced
in Section~\ref{subsec:lifecycle}. Newly synthesized policies are trained from
scratch against their corresponding rewards, while surviving policies retain
fixed parameters across subsequent generations.

\begin{algorithm}[h]
\caption{Habitats evolutionary lifecycle}
\label{alg:habitats}
\begin{algorithmic}[1]
\REQUIRE Initial population $\mathcal{P}_1$, number of evaluated generations
$T$, environment description $E_{\mathrm{desc}}$, grounding threshold
$\lambda$, and numerical regularizer $\epsilon>0$.
\ENSURE Final evaluated population $\mathcal{P}_T$, final survivor set
$\mathcal{V}_T$, per-generation metrics, and evolutionary traces.
\FOR{$t=1,\ldots,T$}
\STATE Train each newly initialized policy in $\mathcal{P}_t$ against its
corresponding reward; keep inherited survivor policies fixed.
\STATE Evaluate every policy against every population reward to construct
$\mathbf{F}_t$.
\STATE Evaluate the random-policy baseline under every population reward.
\STATE Independently normalize each column of $\mathbf{F}_t$ to obtain
$\tilde{\mathbf{F}}_t$.
\STATE Compute $\mathrm{SQ}_t$, $\mathrm{BD}_t$, the grounded set
$\mathcal{G}_t$, and $\mathrm{GC}_t$.
\STATE Select the survivor set $\mathcal{V}_t$ using niche dominance and
the random-policy baseline.
\IF{$t<T$}
\STATE The Planner selects $K_t$ and produces parent--directive pairs
$\{(P_k,h_k)\}_{k=1}^{K_t}$.
\STATE Initialize the valid offspring set
$\mathcal{O}_t\leftarrow\emptyset$.
\FOR{$k=1,\ldots,K_t$}
\STATE Initialize the attempt counter $a_k\leftarrow 0$ and validity flag
$v_k\leftarrow\mathrm{false}$.
\WHILE{$a_k<4 \land v_k=\mathrm{false}$}
\STATE Synthesize a candidate reward
$R_{\mathrm{child},k}$ using the Mutator.
\STATE Validate the candidate using static syntax checking and sandboxed execution.
\STATE $a_k\leftarrow a_k+1$.
\IF{$R_{\mathrm{child},k}$ is valid}
\STATE $v_k\leftarrow\mathrm{true}$.
\ENDIF
\ENDWHILE
\IF{$v_k=\mathrm{true}$}
\STATE Pair $R_{\mathrm{child},k}$ with a newly initialized policy and add
the pair to $\mathcal{O}_t$.
\ELSE
\STATE Discard offspring slot $k$.
\ENDIF
\ENDFOR
\STATE Update the population:
$\mathcal{P}_{t+1}\leftarrow\mathcal{V}_t\cup\mathcal{O}_t$.
\ENDIF
\ENDFOR
\STATE Compute
$\mathrm{GC}_{\mathrm{AUC}}\leftarrow
\sum_{t=1}^{T}\mathrm{GC}_t$.
\RETURN $\mathcal{P}_T$, $\mathcal{V}_T$, and the recorded metrics.
\end{algorithmic}
\end{algorithm}

%%%%%%%%%%%%%%%%%%%%%%%%%%%%%%%%%%%%%%%%%%%%%%%%%%%%%%%%%%%%%%%%
% New Appendix
%%%%%%%%%%%%%%%%%%%%%%%%%%%%%%%%%%%%%%%%%%%%%%%%%%%%%%%%%%%%%%%%
\newpage
\section{Environment Specifications}
\label{app:environment_specs}

As defined in Eq.~\eqref{eq:reward_contract}, every synthesized objective is
implemented as an executable function with the common interface
\begin{equation}
R_i(s,a,s',\mathrm{info})\longrightarrow\mathbb{R}.
\end{equation}
Here, $\mathrm{info}$ abstracts over auxiliary implementation inputs, including
the environment adapter and terminal flag; the concrete Python signatures are
provided in Appendix~\ref{app:prompts}.

Table~\ref{tab:env_specs} summarizes the environment spaces, policy-optimization
algorithms, and evolutionary settings. Native task rewards are not used for
policy optimization.

\begin{table*}[h]
\centering
\caption{Environment specifications and evolutionary settings.}
\label{tab:env_specs}
\small
\setlength{\tabcolsep}{4pt}
\renewcommand{\arraystretch}{1.1}

\begin{tabular}{@{}
p{0.20\textwidth}
p{0.20\textwidth}
p{0.17\textwidth}
p{0.17\textwidth}
p{0.17\textwidth}
@{}}
\toprule
\textbf{Property}
& \textbf{Card Sandbox}
& \textbf{Cart-Pole}
& \textbf{Hopper}
& \textbf{Cheetah}
\tabularnewline
\midrule

Observation space
& Discrete $|\mathcal{S}|=6{,}561$
& Continuous $\mathbb{R}^{5}$
& Continuous $\mathbb{R}^{15}$
& Continuous $\mathbb{R}^{17}$
\tabularnewline

Action space
& Discrete, $|\mathcal{A}|=9$
& Continuous $\mathbb{R}^{1}$
& Continuous $\mathbb{R}^{4}$
& Continuous $\mathbb{R}^{6}$
\tabularnewline

Policy optimization
& Tabular Q-learning
& PPO
& PPO
& PPO
\tabularnewline

Train episode horizon
& $500$
& $1{,}000$
& $1{,}000$
& $1{,}000$
\tabularnewline

Initial population $N_\text{init}$
& $3$
& $4$
& $4$
& $4$
\tabularnewline

Generations $T$
& $15$
& $6$
& $6$
& $6$
\tabularnewline
\bottomrule
\end{tabular}
\end{table*}

\paragraph{Card Sandbox.}
The Card Sandbox is implemented as a custom
single-player card environment. It operates over a 32-card deck
containing four copies of each card value from $1$ to $8$. The agent maintains a hand with a maximum size of one and interacts with three table slots.

The environment has nine discrete actions:
\[
\mathcal{A}
=
\{
\textsc{pass},
\textsc{draw},
\textsc{play}_{0},
\textsc{play}_{1},
\textsc{play}_{2},
\textsc{take}_{0},
\textsc{take}_{1},
\textsc{take}_{2},
\textsc{discard}
\}.
\]
Action $0$ passes, action $1$ draws from the deck, actions $2$--$4$ play the
card in hand to table slots $0$--$2$, actions $5$--$7$ retrieve cards from
table slots $0$--$2$, and action $8$ discards the card in hand. 

The observation is represented as a discrete scalar with
$9^{4}=6{,}561$ possible values. Each of the hand and three table positions
contains either $0$ for an empty position or a card value in
$\{1,\ldots,8\}$. The state index is encoded deterministically as
\[
s
=
729\,h
+
81\,x_0
+
9\,x_1
+
x_2,
\]
where $h$ is the card value in hand and $x_0,x_1,x_2$ are the values in the
three table slots.

Synthesized reward functions receive the encoded pre- and post-transition
states together with structured metadata in \texttt{info}. This metadata
exposes the decoded state components, raw hand and table arrays, remaining deck
contents, action names, and the current step count. The Card Sandbox does not
provide a predefined task reward during policy optimization; all learned
behavior is induced by the synthesized reward programs.

\paragraph{MuJoCo control environments.}
The continuous-control experiments use the DeepMind Control Suite \citep{tassa2018deepmind} with the MuJoCo physics engine. We evaluate the domain--task pairs \texttt{cartpole/balance},
\texttt{hopper/hop}, and \texttt{cheetah/run}, covering balancing and
locomotion behaviors. These task identifiers are used only to instantiate the underlying control environments; their native task rewards are discarded and are not used for policy optimization or cross-fitness evaluation. Observations and bounded continuous actions are represented using single-precision \texttt{float32} values.

For \texttt{cartpole}, the policy receives a five-dimensional
observation and produces a one-dimensional action bounded to $[-1,1]$. The
metadata supplied to synthesized reward functions includes
\texttt{cart\_position}, \texttt{pole\_angle\_cosine}, and
\texttt{pole\_angular\_velocity}.

For \texttt{hopper}, the policy receives a 15-dimensional observation and
produces a four-dimensional action bounded component-wise to $[-1,1]$. The
metadata includes \texttt{speed}, \texttt{height}, and \texttt{touch}
foot-contact sensor signals.

For \texttt{cheetah}, the policy receives a 17-dimensional observation and
produces a six-dimensional action bounded component-wise to $[-1,1]$. The
metadata includes \texttt{speed}, \texttt{torso\_height}, and
\texttt{torso\_angle}.

MuJoCo episodes terminate after a maximum horizon of $1{,}000$ steps or when
an applicable domain-specific physical termination condition is reached.
Each synthesized reward function
$R_i(s,a,s',\mathrm{info})$ receives the raw pre-transition state, action,
post-transition state, and domain-specific diagnostic metadata. Newly
generated policies are initialized and trained from scratch using PPO, while
surviving policies retain their learned parameters and are not updated in
subsequent generations.

%%%%%%%%%%%%%%%%%%%%%%%%%%%%%%%%%%%%%%%%%%%%%%%%%%%%%%%%%%%%%%%%
% New Appendix
%%%%%%%%%%%%%%%%%%%%%%%%%%%%%%%%%%%%%%%%%%%%%%%%%%%%%%%%%%%%%%%%
\newpage
\section{Evaluation and Metric Details}
\label{app:evaluation}

\subsection{Cross-fitness evaluation}
For a population $\mathcal{P}_t$ of size $N_t$, each policy $\pi_i$ is
evaluated under each reward function $R_j$. Let $G_{i,j}^{(e)}$ denote the
return obtained during evaluation rollout
$e\in\{1,\ldots,E_{\mathrm{eval}}\}$. For an episode containing $H_e$
transitions, the return is the undiscounted cumulative synthesized reward:
\begin{equation}
G_{i,j}^{(e)}
=
\sum_{\tau=0}^{H_e-1}
R_j
\left(
s_\tau,
a_\tau,
s_{\tau+1},
\mathrm{info}_\tau
\right).
\label{eq:app_episode_return}
\end{equation}
The empirical cross-fitness entry is the mean return across evaluation
rollouts:
\begin{equation}
\mathbf{F}_t[i,j]
=
\frac{1}{E_{\mathrm{eval}}}
\sum_{e=1}^{E_{\mathrm{eval}}}
G_{i,j}^{(e)}.
\label{eq:app_crossfitness}
\end{equation}

Although policy optimization internally uses a discount factor $\gamma$,
cross-fitness evaluation does not discount rewards. Evaluation episodes end
when the environment emits a terminal timestep or when the evaluation horizon
is reached. The evaluation horizon is $H=100$ steps for the Card Sandbox and
$H=1{,}000$ steps for the MuJoCo environments. When an episode reaches the
maximum horizon without naturally terminating, its return is the sum of the
rewards observed up to that horizon; no value-function bootstrap estimate is
added.

\subsection{Cross-fitness normalization.}
Synthesized rewards may differ substantially in scale and range. To make
returns comparable across reward functions, each column of the cross-fitness
matrix is independently min--max normalized:
\begin{equation}
\tilde{\mathbf{F}}_t[i,j]
=
\frac{
\mathbf{F}_t[i,j]
-
\min_{1\leq k\leq N_t}\mathbf{F}_t[k,j]
}{
\max_{1\leq k\leq N_t}\mathbf{F}_t[k,j]
-
\min_{1\leq k\leq N_t}\mathbf{F}_t[k,j]
+
\epsilon_0
}.
\label{eq:app_normalization}
\end{equation}
We use $\epsilon_0=10^{-8}$ for numerical stability. If all policies obtain
the same return under a reward function, the corresponding normalized column
is zero.

\subsection{Specialist Quality}
Specialist Quality measures how effectively policies optimize their paired
objectives relative to the other policies in the current population. It is
defined as the mean normalized diagonal of the cross-fitness matrix:
\begin{equation}
\mathrm{SQ}_t
=
\frac{1}{N_t}
\sum_{i=1}^{N_t}
\tilde{\mathbf{F}}_t[i,i].
\label{eq:app_sq}
\end{equation}

\subsection{Behavioral Diversity}
Let
\begin{equation}
\mathbf{f}_t[i]
=
\tilde{\mathbf{F}}_t[i,:]
\in
[0,1]^{N_t}
\label{eq:app_profile}
\end{equation}
denote the within-generation cross-fitness profile of policy $\pi_i$. Before
computing Behavioral Diversity, zero-norm profiles are excluded and the
remaining profiles are reindexed. Let $N_t^{\mathrm{BD}}$ denote the number of
remaining profiles. No zero-norm profiles occurred in the reported experiments.

Behavioral Diversity is the mean pairwise cosine distance between the remaining
profiles:
\begin{equation}
\mathrm{BD}_t
=
\frac{2}{
N_t^{\mathrm{BD}}
\left(N_t^{\mathrm{BD}}-1\right)
}
\sum_{1\leq i<j\leq N_t^{\mathrm{BD}}}
\left(
1-
\frac{
\mathbf{f}_t[i]\cdot\mathbf{f}_t[j]
}{
\|\mathbf{f}_t[i]\|_2
\|\mathbf{f}_t[j]\|_2
+
\epsilon_0
}
\right).
\label{eq:app_bd}
\end{equation}
If $N_t^{\mathrm{BD}}<2$, we define $\mathrm{BD}_t=0$.

\subsection{Grounding criterion.}
For each reward function $R_i$, we independently evaluate a uniformly random
policy and estimate its mean and standard deviation of returns,
$\mu_i^{\mathrm{rand}}$ and $\sigma_i^{\mathrm{rand}}$. The learning-progress
score of the paired trained policy is
\begin{equation}
L_i
=
\frac{
\mathbf{F}_t[i,i]
-
\mu_i^{\mathrm{rand}}
}{
\sigma_i^{\mathrm{rand}}
+
\epsilon_0
}.
\label{eq:app_learning_progress}
\end{equation}
A reward is considered grounded when it executes successfully and satisfies
\begin{equation}
L_i>\lambda.
\end{equation}
We use $\lambda=0.5$ and denote the set of grounded individuals at generation
$t$ by $\mathcal{G}_t$.

\subsection{Grounded Behavioral Profile Coverage}
Profiles from different generations are not directly comparable because the
population rewards, and therefore the profile coordinates, change throughout
evolution. We consequently compute coverage separately within each generation.

For every grounded individual $i\in\mathcal{G}_t$, we use its normalized
cross-fitness profile
\begin{equation}
\mathbf{f}_t[i]
=
\tilde{\mathbf{F}}_t[i,:].
\end{equation}
Each coordinate is rounded to one decimal place, and duplicate quantized
profiles are removed. The resulting per-generation profile set is
\begin{equation}
\mathcal{C}_t
=
\left\{
\operatorname{round}\!\left(\mathbf{f}_t[i],1\right)
:
i\in\mathcal{G}_t
\right\},
\label{eq:app_profile_set}
\end{equation}
where rounding is applied element-wise. Grounded Behavioral Profile Coverage
at generation $t$ is
\begin{equation}
\mathrm{GC}_t
=
|\mathcal{C}_t|.
\label{eq:app_coverage}
\end{equation}

We report final-generation coverage and coverage accumulated across the
evolutionary trajectory:
\begin{equation}
\mathrm{GC}_{\mathrm{final}}
=
\mathrm{GC}_{T},
\qquad
\mathrm{GC}_{\mathrm{AUC}}
=
\sum_{t=1}^{T}\mathrm{GC}_t.
\label{eq:app_coverage_summary}
\end{equation}
The accumulated measure counts grounded profile instances across generations.

%%%%%%%%%%%%%%%%%%%%%%%%%%%%%%%%%%%%%%%%%%%%%%%%%%%%%%%%%%%%%%%%
% New Appendix
%%%%%%%%%%%%%%%%%%%%%%%%%%%%%%%%%%%%%%%%%%%%%%%%%%%%%%%%%%%%%%%%
\newpage
\section{System Prompts and Evolutionary Operator Templates}
\label{app:prompts}

This appendix documents the system instructions and context templates used by the decoupled two-stage Planner--Mutator evolutionary architecture. Offspring synthesis operates
in two phases: (1) high-level population planning through the Offspring Planner
$\mathcal{M}_{\mathrm{plan}}$, which inspects cross-fitness matrices and
population-health metrics to produce structured breeding directives; and
(2) low-level reward synthesis through the Mutator
$\mathcal{M}_{\mathrm{mutate}}$, which translates these directives and any
selected parent code into executable Python reward functions.

%%%%%%%%%%%%%%%%%%%%%%%%%%%%%%%%%%%%%%%%%%%%%%%%%%%%%%%%%%%%%%%%
% D.1 Offspring Planner Prompts
%%%%%%%%%%%%%%%%%%%%%%%%%%%%%%%%%%%%%%%%%%%%%%%%%%%%%%%%%%%%%%%%
\subsection{Offspring Planner Prompts}

The Offspring Planner acts as an autonomous evolutionary-algorithm controller.
It evaluates population-level health and cross-fitness information to determine
whether the population requires exploration, such as seeding new behavioral
niches or combining distant policies, or exploitation, such as refining
high-performing specialists.

\subsubsection{Planner System Instruction}

The following system instruction is supplied to
$\mathcal{M}_{\mathrm{plan}}$. It defines the Planner's reasoning protocol,
steering rules based on Specialist Quality $\mathrm{SQ}_t$ and Behavioral
Diversity $\mathrm{BD}_t$, and structured JSON output schema.

\begin{quote}
\small
\ttfamily
\raggedright

You are an expert RL researcher and evolutionary algorithm controller.\\
Your task is to plan the next generation of reward functions in an
evolutionary run.\\
You will be given the current generation's agents, their evolved reward codes,
their performance fitness grid, the agents selected as survivors, and a
Population Health Report containing quality and diversity metrics.

\medskip

You must decide how many new offspring to generate and how to generate them.\\
You can choose to mutate existing survivors or generate entirely new
individuals without parents to explore radically new areas of the search
space.

\medskip

For each offspring:
\begin{enumerate}[leftmargin=16pt,topsep=2pt,itemsep=1pt]
    \item Select one to three parents from the survivors, or leave
    \texttt{parent\_ids} empty (\texttt{[]}) to generate an entirely new
    individual.

    \item Provide reasoning for the parent selection, or for starting fresh,
    together with the mutation or generation strategy.

    \item Provide a hint or direction for the mutator/generator. Examples
    include ``combine the exploration strategy of
    \texttt{agent\_1} with the exploitation strategy of
    \texttt{agent\_2},'' ``make a small change to
    \texttt{agent\_3}'s weights to improve performance,'' or, when generating
    a new individual, ``focus on maximizing exploration.''
\end{enumerate}

Population Health Report Metrics:
\begin{itemize}[leftmargin=16pt,topsep=2pt,itemsep=1pt]
    \item \textbf{Specialist Quality:} Measures how well policies perform on
    their own target rewards on average, from 0.0 to 1.0.
    \begin{itemize}[leftmargin=16pt,topsep=1pt,itemsep=1pt]
        \item High (\textasciitilde 1.0) means agents are successfully
        learning and specializing.
        \item Low ($<0.5$) means agents are struggling to learn or one
        dominant policy is winning across objectives.
    \end{itemize}

    \item \textbf{Behavioral Diversity:} Measures how different the policies'
    performance profiles are from one another, from 0.0 to 1.0.
    \begin{itemize}[leftmargin=16pt,topsep=1pt,itemsep=1pt]
        \item High (\textasciitilde 1.0) means policies have discovered very
        different strategies.
        \item Low ($<0.3$) means the population is homogenizing, risking
        stagnation.
    \end{itemize}
\end{itemize}

Use the Population Health Report to steer evolution as follows:
\begin{itemize}[leftmargin=16pt,topsep=2pt,itemsep=1pt]
    \item \textbf{If Specialist Quality is LOW and Behavioral Diversity is
    HIGH:} The population is in chaos. Rewards may be too different or too
    difficult. Prioritize EXPLOITATION by proposing simpler, more incremental
    offspring rewards.

    \item \textbf{If Specialist Quality is HIGH and Behavioral Diversity is
    LOW:} The population is stagnating. Prioritize EXPLORATION by proposing
    radical mutations, combining substantially different parents, or
    introducing new structural concepts.

    \item \textbf{If both are HIGH:} The population is healthy. Maintain a
    balance between incremental improvements and continued exploration.
\end{itemize}

Constraints:
\begin{itemize}[leftmargin=16pt,topsep=2pt,itemsep=1pt]
    \item You must select parents only from the list of survivors, unless
    generating a new individual, in which case leave
    \texttt{parent\_ids} empty.
\end{itemize}

You must output the plan as JSON matching the following schema:

\medskip

\{\\
\hspace*{1em}"offspring\_plans": [\\
\hspace*{2em}\{\\
\hspace*{3em}"parent\_ids": ["agent\_id\_1", "agent\_id\_2"],\\
\hspace*{3em}"reasoning": "Description of why these parents were chosen...",\\
\hspace*{3em}"mutation\_hint": "Specific instructions for the mutator..."\\
\hspace*{2em}\}\\
\hspace*{1em}]\\
\}

\end{quote}

\subsubsection{Planner User Context Template}

At generation $t$, $\mathcal{M}_{\mathrm{plan}}$ receives the full population
state, cross-fitness matrix, and quality-diversity metrics through the following
template:

\begin{quote}
\small
\ttfamily
\raggedright

Environment Description:\\
\{env\_desc\}

\medskip

Current Generation: \{generation\}\\
Number of Survivors: \{num\_survivors\}

\medskip

--- POPULATION HEALTH REPORT ---\\
Specialist Quality: \{specialist\_quality\}\\
Behavioral Diversity: \{behavioral\_diversity\}

\medskip

Survivors (Selected for next generation):\\
\{survivors\}

\medskip

All Agents in this Generation:\\
\{all\_agents\}

\medskip

Fitness Grid (Row $i$ is Policy $i$, Column $j$ is Reward $j$):\\
\{fitness\_grid\}

\medskip

Analyze the data and generate the offspring plans.

\end{quote}

\subsection{Reward Synthesizer and Mutator Prompts}

The Mutator translates Planner directives and selected parent source code into
executable Python functions. To reduce programmatic stagnation and code
duplication, it is also conditioned on a rolling window of previously generated
reward functions, denoted by $\mathcal{H}_{\mathrm{forbidden}}$.

\subsubsection{Mutator System Instruction}

The system instruction enforces reward simplicity, valid Python syntax, and the
required function contract.

\begin{quote}
\small
\ttfamily
\raggedright

You are an expert RL researcher and evolutionary operator.\\
Your task is to design, mutate, and improve Python reward functions for a
reinforcement learning agent to elicit interesting and diverse behaviors in a
PLAYGROUND environment.

\medskip

The original game-theoretic objectives, such as winning or reaching a
predefined goal, are COMPLETELY DISREGARDED.\\
The environment is a playground for exploration. 

\medskip

You must output a new reward function in Python. The output MUST be a valid
Python code block beginning with
\texttt{\textasciigrave\textasciigrave\textasciigrave python} and ending with
\texttt{\textasciigrave\textasciigrave\textasciigrave}. It must define the
following function:

\medskip

\texttt{def compute\_reward(env, state, action, next\_state, done, info}\\
\texttt{\hspace*{1em}) -> float:}

\medskip

Inside this function, you may decode states using environment-specific methods.\\
Do not use external libraries other than standard Python libraries, such as
\texttt{math}.

\medskip

The function must return a single finite Python \texttt{float} representing
the synthesized reward for the current transition. The function cannot modify
the environment's termination signal.

Explain the strategy and behavioral objective in comments inside the code.

\medskip

CRITICAL REQUIREMENTS FOR REWARD DESIGN:
\begin{enumerate}[leftmargin=16pt,topsep=2pt,itemsep=1pt]
    \item Keep the reward function as SIMPLE and interpretable as possible.
    Avoid unnecessary complexity.

    \item Define a clear and simple behavioral objective.

    \item Prefer simple, sparse, or discrete rewards, such as
    \texttt{+1.0} for goal progress, \texttt{-1.0} for invalid actions, and
    \texttt{0.0} otherwise, when sufficient.

    \item Use continuous reward shaping only when necessary to guide the agent
    toward the objective.

    \item Avoid combining many reward components unless essential.

    \item Do NOT define helper functions inside
    \texttt{compute\_reward}. Implement all logic inline or use existing
    environment methods, such as \texttt{env.decode}, to maximize performance.
\end{enumerate}

\end{quote}

\subsubsection{Parent Mutation and Combination Template}

When mutating surviving parents under Planner guidance,
$\mathcal{M}_{\mathrm{mutate}}$ receives the parent source code, Planner
directive, and historical forbidden reward programs through the following
template:

\begin{quote}
\small
\ttfamily
\raggedright

Environment Description:\\
\{env\_desc\}

\medskip

Parent Reward Functions from the previous generation
(use these to mutate or combine):\\
\{parent\_codes\}

\medskip

Previously generated reward functions in this run
(DO NOT duplicate these):\\
\{forbidden\_codes\}

\medskip

Generate a NEW mutated reward function that differs from the forbidden
functions while inheriting or modifying strategies from the selected parents.

\medskip

Mutation Instruction/Hint:\\
\{mutation\_hint\}

\end{quote}

%%%%%%%%%%%%%%%%%%%%%%%%%%%%%%%%%%%%%%%%%%%%%%%%%%%%%%%%%%%%%%%%
% D.3 Environment Description Prompts
%%%%%%%%%%%%%%%%%%%%%%%%%%%%%%%%%%%%%%%%%%%%%%%%%%%%%%%%%%%%%%%%
\subsection{Environment Interface Description Prompts
(\texttt{\{env\_desc\}})}

Whenever the Planner or Mutator is called, the \texttt{\{env\_desc\}}
placeholder is populated with the domain-specific environment specification.
This description provides the LLM with the observation structure, action space,
available helper methods, action bounds, and reward-function signature.

\subsubsection{Discrete Card Sandbox Description}

\begin{quote}
\small
\ttfamily
\raggedright

Environment: Playing Card Playground (single-player)

\medskip

State Space: \texttt{Discrete(6561)} representing the game state.

\medskip

You can decode the state integer using \texttt{env.decode(state)}, which
returns a dictionary:

\medskip

\{\\
\hspace*{1em}"hand": int (0--8, 0 = empty),\\
\hspace*{1em}"table": list of 3 ints (0--8, 0 = empty)\\
\}

\medskip

Action Space: \texttt{Discrete(9)}\\
\hspace*{1em}0: Pass\\
\hspace*{1em}1: Draw (deck to hand)\\
\hspace*{1em}2--4: Play hand to table slots 0--2\\
\hspace*{1em}5--7: Take from table slots 0--2 to hand\\
\hspace*{1em}8: Discard hand

\medskip

This is a PLAYGROUND. There is no predetermined objective, winning condition,
or built-in task reward.

\medskip

Goal: Evolve a SIMPLE and interesting objective, effectively creating a
mini-game for the agent to play in this environment. Examples include arranging
cards on the table in patterns, matching values, hoarding cards, or cycling
cards between the hand and table.

\end{quote}

\subsubsection{Continuous Cart-Pole Description}

\begin{quote}
\small
\ttfamily
\raggedright

Environment: MuJoCo Cart-Pole Playground (Continuous)

\medskip

Observation Space: \texttt{np.ndarray} of shape \texttt{(5,)} representing
the flattened physics state:

\medskip

\hspace*{1em}Index 0: \texttt{cart\_position} (float) -- horizontal position
of the cart on the slider\\
\hspace*{1em}Index 1: \texttt{cos\_pole\_angle} (float) -- cosine of the pole
angle (1.0 = upright, -1.0 = hanging down)\\
\hspace*{1em}Index 2: \texttt{sin\_pole\_angle} (float) -- sine of the pole
angle\\
\hspace*{1em}Index 3: \texttt{cart\_velocity} (float) -- horizontal velocity
of the cart\\
\hspace*{1em}Index 4: \texttt{pole\_angular\_velocity} (float) -- angular
velocity of the pole

\medskip

You can decode \texttt{state} and \texttt{next\_state} arrays using
\texttt{env.decode(state)}, which returns a dictionary:

\medskip

\{\\
\hspace*{1em}"cart\_position": float,\\
\hspace*{1em}"cos\_pole\_angle": float,\\
\hspace*{1em}"sin\_pole\_angle": float,\\
\hspace*{1em}"cart\_velocity": float,\\
\hspace*{1em}"pole\_angular\_velocity": float\\
\}

\medskip

Action Space: \texttt{Continuous(1)} -- an \texttt{np.ndarray} of shape
\texttt{(1,)} in \texttt{[-1.0, 1.0]}, representing the horizontal force
applied to the cart.

\medskip

Convenience properties available on \texttt{env}:\\
\hspace*{1em}\texttt{env.cart\_position} -> float\\
\hspace*{1em}\texttt{env.pole\_angle\_cosine} -> float
(1.0 = upright)\\
\hspace*{1em}\texttt{env.pole\_angle\_sine} -> float\\
\hspace*{1em}\texttt{env.cart\_velocity} -> float\\
\hspace*{1em}\texttt{env.pole\_angular\_velocity} -> float\\
\hspace*{1em}\texttt{env.get\_env()} -> underlying DM Control environment for
advanced physics access

\medskip

Function signature:

\medskip

\texttt{def compute\_reward(env: MujocoEnv, state: np.ndarray,}\\
\texttt{\hspace*{1em}action: np.ndarray, next\_state: np.ndarray,}\\
\texttt{\hspace*{1em}done: bool, info: dict)}\\
\texttt{\hspace*{1em}-> float:}

\medskip

This is a PLAYGROUND. There is no predetermined objective, winning condition,
or built-in task reward.

\medskip

Goal: Evolve a SIMPLE and interesting objective for the agent to explore in
this environment. Examples include balancing the pole upright, swinging the
pole, moving the cart to a target position while balancing, or minimizing cart
movement.

\end{quote}

\subsubsection{Continuous Hopper Description}

\begin{quote}
\small
\ttfamily
\raggedright

Environment: MuJoCo Hopper Playground (Continuous)

\medskip

Observation Space: \texttt{np.ndarray} of shape \texttt{(15,)} representing
the flattened physics state.

\medskip

Positions (indices 0--5):\\
\hspace*{1em}Index 0: \texttt{rootz} (float) -- vertical position (height) of
the torso\\
\hspace*{1em}Index 1: \texttt{rooty} (float) -- angle of the torso
(orientation/tilt)\\
\hspace*{1em}Index 2: \texttt{waist} (float) -- waist joint angle connecting
the torso to the pelvis\\
\hspace*{1em}Index 3: \texttt{hip} (float) -- hip joint angle connecting the
pelvis to the thigh\\
\hspace*{1em}Index 4: \texttt{knee} (float) -- knee joint angle connecting the
thigh to the calf\\
\hspace*{1em}Index 5: \texttt{ankle} (float) -- ankle joint angle connecting
the calf to the foot

\medskip

Velocities (indices 6--12):\\
\hspace*{1em}Index 6: \texttt{rootx\_vel} (float) -- horizontal velocity
(approximately forward speed)\\
\hspace*{1em}Index 7: \texttt{rootz\_vel} (float) -- vertical velocity\\
\hspace*{1em}Index 8: \texttt{rooty\_vel} (float) -- angular velocity of the
torso\\
\hspace*{1em}Index 9: \texttt{waist\_vel} (float) -- waist joint velocity\\
\hspace*{1em}Index 10: \texttt{hip\_vel} (float) -- hip joint velocity\\
\hspace*{1em}Index 11: \texttt{knee\_vel} (float) -- knee joint velocity\\
\hspace*{1em}Index 12: \texttt{ankle\_vel} (float) -- ankle joint velocity

\medskip

Touch sensors (indices 13--14):\\
\hspace*{1em}Index 13: \texttt{touch\_toe} (float) -- \texttt{log1p} of the
toe touch-sensor signal, where 0.0 indicates no contact\\
\hspace*{1em}Index 14: \texttt{touch\_heel} (float) -- \texttt{log1p} of the
heel touch-sensor signal, where 0.0 indicates no contact

\medskip

Horizontal position (\texttt{rootx}) is excluded from the observation to
maintain translational invariance.

\medskip

You can decode \texttt{state} and \texttt{next\_state} arrays using
\texttt{env.decode(state)}, which returns a dictionary:

\medskip

\{\\
\hspace*{1em}"rootz": float, "rooty": float,\\
\hspace*{1em}"waist": float, "hip": float, "knee": float, "ankle": float,\\
\hspace*{1em}"rootx\_vel": float, "rootz\_vel": float,
"rooty\_vel": float,\\
\hspace*{1em}"waist\_vel": float, "hip\_vel": float,
"knee\_vel": float, "ankle\_vel": float,\\
\hspace*{1em}"touch\_toe": float, "touch\_heel": float\\
\}

\medskip

Action Space: \texttt{Continuous(4)} -- an \texttt{np.ndarray} of shape
\texttt{(4,)} in \texttt{[-1.0, 1.0]}, representing torques applied to four
joints:

\medskip

\hspace*{1em}Index 0: waist torque (gear ratio 30)\\
\hspace*{1em}Index 1: hip torque (gear ratio 40)\\
\hspace*{1em}Index 2: knee torque (gear ratio 30)\\
\hspace*{1em}Index 3: ankle torque (gear ratio 10)

\medskip

Convenience properties available on \texttt{env}:\\
\hspace*{1em}\texttt{env.height} -> float (height of the torso relative to the
foot; standing threshold approximately 0.6)\\
\hspace*{1em}\texttt{env.speed} -> float (horizontal speed of the torso)\\
\hspace*{1em}\texttt{env.touch} -> \texttt{tuple[float, float]}
(\texttt{touch\_toe}, \texttt{touch\_heel})\\
\hspace*{1em}\texttt{env.get\_env()} -> underlying DM Control environment for
advanced physics access

\medskip

Function signature:

\medskip

\texttt{def compute\_reward(env: MujocoEnv, state: np.ndarray,}\\
\texttt{\hspace*{1em}action: np.ndarray, next\_state: np.ndarray,}\\
\texttt{\hspace*{1em}done: bool, info: dict)}\\
\texttt{\hspace*{1em}-> float:}

\medskip

This is a PLAYGROUND. There is no predetermined objective, winning condition,
or built-in task reward.

\medskip

Goal: Evolve a SIMPLE and interesting objective for the agent to explore in
this playground. Examples include standing upright and balancing, hopping
forward, hopping as high as possible, balancing on one toe, minimizing energy
use while standing, or achieving a target speed.

\end{quote}

\subsubsection{Continuous Cheetah Description}

\begin{quote}
\small
\ttfamily
\raggedright

Environment: MuJoCo Cheetah Playground (Continuous)

\medskip

Observation Space: \texttt{np.ndarray} of shape \texttt{(17,)} representing
the flattened physics state.

\medskip

Positions (indices 0--7):\\
\hspace*{1em}Index 0: \texttt{rootz} (float) -- vertical position (height) of
the torso\\
\hspace*{1em}Index 1: \texttt{rooty} (float) -- angle of the torso
(orientation)\\
\hspace*{1em}Index 2: \texttt{bthigh} (float) -- back-thigh joint angle\\
\hspace*{1em}Index 3: \texttt{bshin} (float) -- back-shin joint angle\\
\hspace*{1em}Index 4: \texttt{bfoot} (float) -- back-foot joint angle\\
\hspace*{1em}Index 5: \texttt{fthigh} (float) -- front-thigh joint angle\\
\hspace*{1em}Index 6: \texttt{fshin} (float) -- front-shin joint angle\\
\hspace*{1em}Index 7: \texttt{ffoot} (float) -- front-foot joint angle

\medskip

Velocities (indices 8--16):\\
\hspace*{1em}Index 8: \texttt{rootx\_vel} (float) -- horizontal velocity
(approximately forward speed)\\
\hspace*{1em}Index 9: \texttt{rootz\_vel} (float) -- vertical velocity\\
\hspace*{1em}Index 10: \texttt{rooty\_vel} (float) -- angular velocity of the
torso\\
\hspace*{1em}Index 11: \texttt{bthigh\_vel} (float) -- back-thigh joint
velocity\\
\hspace*{1em}Index 12: \texttt{bshin\_vel} (float) -- back-shin joint
velocity\\
\hspace*{1em}Index 13: \texttt{bfoot\_vel} (float) -- back-foot joint
velocity\\
\hspace*{1em}Index 14: \texttt{fthigh\_vel} (float) -- front-thigh joint
velocity\\
\hspace*{1em}Index 15: \texttt{fshin\_vel} (float) -- front-shin joint
velocity\\
\hspace*{1em}Index 16: \texttt{ffoot\_vel} (float) -- front-foot joint
velocity

\medskip

Horizontal position (\texttt{rootx}) is excluded from the observation to
maintain translational invariance.

\medskip

You can decode \texttt{state} and \texttt{next\_state} arrays using
\texttt{env.decode(state)}, which returns a dictionary:

\medskip

\{\\
\hspace*{1em}"rootz": float, "rooty": float,\\
\hspace*{1em}"bthigh": float, "bshin": float, "bfoot": float,\\
\hspace*{1em}"fthigh": float, "fshin": float, "ffoot": float,\\
\hspace*{1em}"rootx\_vel": float, "rootz\_vel": float,
"rooty\_vel": float,\\
\hspace*{1em}"bthigh\_vel": float, "bshin\_vel": float,
"bfoot\_vel": float,\\
\hspace*{1em}"fthigh\_vel": float, "fshin\_vel": float,
"ffoot\_vel": float\\
\}

\medskip

Action Space: \texttt{Continuous(6)} -- an \texttt{np.ndarray} of shape
\texttt{(6,)} in \texttt{[-1.0, 1.0]}, representing torques applied to six
joints:

\medskip

\hspace*{1em}Index 0: \texttt{bthigh} (back-thigh torque)\\
\hspace*{1em}Index 1: \texttt{bshin} (back-shin torque)\\
\hspace*{1em}Index 2: \texttt{bfoot} (back-foot torque)\\
\hspace*{1em}Index 3: \texttt{fthigh} (front-thigh torque)\\
\hspace*{1em}Index 4: \texttt{fshin} (front-shin torque)\\
\hspace*{1em}Index 5: \texttt{ffoot} (front-foot torque)

\medskip

Convenience properties available on \texttt{env}:\\
\hspace*{1em}\texttt{env.speed} -> float (horizontal speed of the torso)\\
\hspace*{1em}\texttt{env.torso\_height} -> float (vertical position of the
torso)\\
\hspace*{1em}\texttt{env.torso\_angle} -> float (orientation angle of the
torso)\\
\hspace*{1em}\texttt{env.get\_env()} -> underlying DM Control environment for
advanced physics access

\medskip

Function signature:

\medskip

\texttt{def compute\_reward(env: MujocoEnv, state: np.ndarray,}\\
\texttt{\hspace*{1em}action: np.ndarray, next\_state: np.ndarray,}\\
\texttt{\hspace*{1em}done: bool, info: dict)}\\
\texttt{\hspace*{1em}-> float:}

\medskip

This is a PLAYGROUND. There is no predetermined objective, winning condition,
or built-in task reward.

\medskip

Goal: Evolve a SIMPLE and interesting objective for the agent to explore in
this playground. Examples include running forward at high speed, running
backward, maintaining a target torso height or posture, performing flips, or
minimizing torque and energy use.

\end{quote}

%%%%%%%%%%%%%%%%%%%%%%%%%%%%%%%%%%%%%%%%%%%%%%%%%%%%%%%%%%%%%%%%
% New Appendix
%%%%%%%%%%%%%%%%%%%%%%%%%%%%%%%%%%%%%%%%%%%%%%%%%%%%%%%%%%%%%%%%
\newpage
\section{Hyperparameters, Implementation, Compute, and Random Seeds}
\label{app:hyperparameters}

\subsection{Policy-optimization hyperparameters.}
Table~\ref{tab:hyperparams} reports the policy-training and evaluation settings across discrete and continuous environments.

\begin{table}[h]
\centering
\caption{Policy-training and evaluation hyperparameters.}
\label{tab:hyperparams}
\small
\setlength{\tabcolsep}{4pt}
\begin{tabular}{
p{0.32\linewidth}
p{0.27\linewidth}
p{0.33\linewidth}}
\toprule
\textbf{Hyperparameter}
& \textbf{Card Sandbox}
& \textbf{Continuous control} \\
\midrule
Policy optimizer
& Tabular Q-learning
& PPO (Brax/JAX) \\

Learning rate
& \(0.05\)
& Cart-Pole/Cheetah: \(3\times10^{-4}\);
  Hopper: \(6\times10^{-4}\) \\

Discount factor \(\gamma\)
& \(0.95\)
& \(0.99\) \\

Exploration or clipping
& \(\epsilon\)-greedy, \(1.0\rightarrow0.01\) decay
& PPO clip ratio \(\epsilon=0.2\) \\

Training budget per new policy
& \(20{,}000\) episodes
& \(100{,}000\) environment steps \\

Evaluation rollouts \(E_{\mathrm{eval}}\)
& \(10\) episodes
& \(10\) episodes \\

Train episode horizon \(H\)
& \(500\) steps
& \(1{,}000\) steps \\

GAE parameter \(\lambda_{\mathrm{gae}}\)
& Not applicable
& \(0.95\) \\

PPO epochs per update
& Not applicable
& \(10\) \\

Minibatch size
& Not applicable
& \(64\) \\
\bottomrule
\end{tabular}
\end{table}

We run these settings using the DM Control library \citep{tassa2018deepmind} and ACME \citep{hoffman2020acme} libraries.

\subsection{Language-model configuration.}

Unless otherwise specified, all evolutionary language-model operations use
\texttt{gemini-3.5-flash}. In the backbone ablation, both the Planner and
Mutator instead use \texttt{gemini-3.8-flash} or
\texttt{gemini-4-argon}, with all other settings held fixed.

\begin{itemize}[leftmargin=12pt,topsep=2pt,itemsep=2pt]
\item \textbf{Decoding settings:} Sampling parameters are not explicitly set
in \texttt{GenerateContentConfig}. The SDK defaults used during the experiments
are temperature $1.0$, top-$p$ $0.95$, and a maximum output length of
$8{,}192$ tokens. No API seed is provided.

\item \textbf{Structured output:} Offspring Planner responses are constrained
to JSON objects adhering to a predefined schema, with
\texttt{offspring\_plans} containing \texttt{parent\_ids},
\texttt{reasoning}, and \texttt{mutation\_hint}. Mutator responses are parsed
to extract executable Python code blocks.

\item \textbf{Retry policy and transient error handling:} The custom
\texttt{LLMClient} wrapper retries transient API failures, network timeouts,
and quota rate limits up to five times using exponential backoff before raising
an exception.

\item \textbf{Sandbox code validation and replacement:} Synthesized reward
functions undergo static AST syntax checking and sandboxed execution before
policy training. If validation fails, the candidate is rejected and the
Mutator is called again, for up to three replacement attempts. If all four
attempts for an offspring slot fail, that slot is discarded. Only validated
candidates proceed to policy training.
\end{itemize}

\subsection{Software, database infrastructure, and compute.}\label{infrastructure}
The Habitats framework is implemented in Python 3.13 using JAX and PyTorch.

\begin{itemize}[leftmargin=12pt,topsep=2pt,itemsep=2pt]
\item \textbf{Distributed Asynchronous Execution:} To minimize wall-clock runtime, policy evaluation rollouts and independent Q-learning training runs are dispatched asynchronously across multi-threaded worker pools (\texttt{ThreadPoolExecutor}) configured with \texttt{max\_workers = 5} for evaluation and up to $10$ workers for parallel environment interaction. They run on a chip with 8 CPUs and 32GB of RAM. On average, a single Q-Learning experiment takes around 1h to run. PPO policy training for continuous MuJoCo domains runs on dedicated NVIDIA H100 GPU host machines that have 8 CPUs and 32GB of RAM. A single PPO experiment takes on average around 2h to complete. 

\item \textbf{State Persistence and Auditing:} Full experimental state (including run metadata, generation summaries, raw and normalized cross-fitness matrices $\mathbf{F}_t$, Specialist Quality, Behavioral Diversity, agent reward code histories, and LLM reasoning traces) is continuously persisted to a relational SQLite database. Policy weights are checkpointed incrementally.

\end{itemize}

\subsection{Random seeds and experimental control.}

To account for stochasticity, all reported results are aggregated across three
independent evolutionary runs initialized with
\begin{equation}
\mathcal{S}_{\mathrm{seeds}}
=
\{11116306,\;56229370,\;67374653\}.
\end{equation}
These seeds initialize the Python and NumPy random-number generators, controlling
environment initialization and tabular Q-learning exploration. No deterministic
seed is provided to the Gemini API, so language-model generations and complete
end-to-end runs are not guaranteed to be exactly reproducible.

%%%%%%%%%%%%%%%%%%%%%%%%%%%%%%%%%%%%%%%%%%%%%%%%%%%%%%%%%%%%%%%%
% New Appendix
%%%%%%%%%%%%%%%%%%%%%%%%%%%%%%%%%%%%%%%%%%%%%%%%%%%%%%%%%%%%%%%%
\newpage
\section{Baseline and Budget-Matching Details}
\label{app:baselines}

To assess the contributions of population-level LLM planning
($\mathcal{M}_{\mathrm{plan}}$) and LLM-guided code mutation
($\mathcal{M}_{\mathrm{mutate}}$), Habitats is evaluated against two
evolutionary baselines: \emph{Independent Generation} and
\emph{Direct Mutation}. All three methods use identical policy-optimization
algorithms, evaluation protocols, environment settings, and survivor-selection
rules. They differ in how offspring reward functions are generated and allocated.

\subsection{Baseline Definitions}

\paragraph{Independent Generation
(\texttt{IndependentOffspringStrategy}).}
At each generation $t$, offspring reward functions are generated independently
from the environment-description prompt,
$\mathcal{M}_{\mathrm{mutate}}(E_{\mathrm{desc}})$, without parent
conditioning, crossover, or Planner directives. Historical duplicate avoidance
through $\mathcal{H}_{\mathrm{forbidden}}$ and niche-dominance survivor
selection remain active, but offspring have no lineage-based dependence on
previous generations.

\paragraph{Direct Mutation
(\texttt{DirectMutationOffspringStrategy}).}
Surviving individuals in $\mathcal{V}_{t}$ are selected in round-robin order
and passed directly to $\mathcal{M}_{\mathrm{mutate}}$ with a generic variation
prompt. This preserves lineage-based reward inheritance and incremental
mutation while removing population-level planning through
$\mathcal{M}_{\mathrm{plan}}$ and structured adaptation based on Specialist
Quality or Behavioral Diversity.

\paragraph{Habitats
(\texttt{HabitatsOffspringStrategy}).}
Habitats uses the complete two-stage architecture. The Offspring Planner
$\mathcal{M}_{\mathrm{plan}}$ inspects the normalized cross-fitness matrix
$\tilde{\mathbf{F}}_{t}$ and the population-health report
$(\mathrm{SQ}_{t},\mathrm{BD}_{t})$. It then produces structured breeding
plans specifying parent selections, crossover targets, and mutation directives,
which are executed by $\mathcal{M}_{\mathrm{mutate}}$.

For both baselines, the number of offspring slots is set to
$K_t=|\mathcal{V}_t|$, allocating one offspring slot per survivor. Thus, before
accounting for offspring slots discarded after failed validation, the target
next-generation population size is $2|\mathcal{V}_t|$. For Habitats, $K_t$ is determined dynamically by the planner.

\subsection{Budget-Matching Protocol}

We exactly match per-policy training and per-pair evaluation budgets, while
total language-model inference and policy-training costs may vary with the
resulting offspring and population sizes.

\paragraph{Language-model inference allocation.}

For $K_t$ offspring slots, each method uses between $K_t$ and $4K_t$ Mutator
calls because every slot permits one initial generation attempt and up to three
replacement attempts. Habitats additionally uses one Planner call per
generation, whereas Independent Generation and Direct Mutation use no Planner
calls. 

\paragraph{Policy-training and evaluation budgets.}
Every newly generated policy receives the same environment-interaction budget
within a given environment. For a population of size $N_t$, cross-fitness
evaluation requires $N_t^2$ policy--reward evaluations per generation. Each
policy $\pi_i$ is evaluated under each reward $R_j$ using the same number of
rollouts and the same episode horizon across all methods. The evaluation horizon
is $H=100$ for the Card Sandbox and $H=1{,}000$ for the MuJoCo environments.

Because population size $N_t$ evolves endogenously, the total number of
environment interactions and the total number of cross-fitness rollouts are not
constrained to be identical across methods.

\paragraph{Invalid candidates.}
Every candidate reward generated by $\mathcal{M}_{\mathrm{mutate}}$ undergoes
Python abstract-syntax-tree parsing and sandboxed execution before policy
training. Each offspring slot permits one initial generation attempt and up to
three replacement attempts. If all four candidates are invalid, the offspring
slot is discarded. The same validation and replacement protocol is used for
all methods.

%%%%%%%%%%%%%%%%%%%%%%%%%%%%%%%%%%%%%%%%%%%%%%%%%%%%%%%%%%%%%%%%
% New Appendix
%%%%%%%%%%%%%%%%%%%%%%%%%%%%%%%%%%%%%%%%%%%%%%%%%%%%%%%%%%%%%%%%
\newpage
\section{Full Results Tables}
\label{app:full_results}

Tables~\ref{tab:full_sq_bd_results} and~\ref{tab:full_gc_results}
report the complete numerical results underlying
Fig.~\ref{fig:fig2}. All entries are reported as mean
$\pm$ standard deviation across three seeds. Within each environment and metric, the highest mean is shown in bold. Mean SQ and mean BD are averaged over the evaluated generations, whereas final SQ, final BD, and final GC are computed on the final evaluated population $\mathcal{P}_T$. $\mathrm{GC}_{\mathrm{AUC}}$ is the sum of per-generation grounded coverage over the evolutionary trajectory. We additionally report equal-weight macro-averages across environments for SQ and BD. We do not aggregate GC across environments because its scale depends on the number of evaluated generations and population size, both of which vary across environments.

\begin{table*}[h]
\centering
\caption{
Complete Specialist Quality (SQ) and Behavioral Diversity (BD) results.
}
\label{tab:full_sq_bd_results}
\tiny
\setlength{\tabcolsep}{6pt}
\begin{tabular}{llcccc}
\toprule
\textbf{Environment}
& \textbf{Method}
& \textbf{Mean SQ}
& \textbf{Final SQ}
& \textbf{Mean BD}
& \textbf{Final BD} \\
\midrule

\multirow{3}{*}{Card Sandbox}
& Independent Generation
& $0.9558 \pm 0.0146$
& $0.9294 \pm 0.0998$
& $0.4725 \pm 0.0754$
& $0.3815 \pm 0.1800$ \\
& Direct Mutation
& $\mathbf{0.9919 \pm 0.0007}$
& $\mathbf{0.9907 \pm 0.0131}$
& $0.4259 \pm 0.2218$
& $0.3906 \pm 0.2662$ \\
& Habitats
& $0.9683 \pm 0.0227$
& $0.9539 \pm 0.0277$
& $\mathbf{0.6710 \pm 0.0289}$
& $\mathbf{0.6676 \pm 0.2036}$ \\
\midrule

\multirow{3}{*}{Cart-Pole}
& Independent Generation
& $0.3591 \pm 0.1613$
& $0.2988 \pm 0.2252$
& $0.1971 \pm 0.1713$
& $0.1672 \pm 0.1631$ \\
& Direct Mutation
& $\mathbf{0.5152 \pm 0.1324}$
& $0.4085 \pm 0.1094$
& $\mathbf{0.3163 \pm 0.1896}$
& $0.2882 \pm 0.1871$ \\
& Habitats
& $0.4162 \pm 0.1309$
& $\mathbf{0.6615 \pm 0.3564}$
& $0.2987 \pm 0.1409$
& $\mathbf{0.4456 \pm 0.0165}$ \\
\midrule

\multirow{3}{*}{Hopper}
& Independent Generation
& $0.7585 \pm 0.1394$
& $0.8000 \pm 0.2000$
& $0.0384 \pm 0.0333$
& $0.0373 \pm 0.0365$ \\
& Direct Mutation
& $\mathbf{0.9462 \pm 0.0932}$
& $\mathbf{1.0000 \pm 0.000}$
& $0.0281 \pm 0.0072$
& $0.0409 \pm 0.0023$ \\
& Habitats
& $0.8163 \pm 0.1624$
& $0.8537 \pm 0.2025$
& $\mathbf{0.2139 \pm 0.1464}$
& $\mathbf{0.1981 \pm 0.1172}$ \\
\midrule

\multirow{3}{*}{Cheetah}
& Independent Generation
& $0.8867 \pm 0.0075$
& $\mathbf{0.9602 \pm 0.0165}$
& $0.0516 \pm 0.0088$
& $0.0349 \pm 0.0099$ \\
& Direct Mutation
& $0.9102 \pm 0.0203$
& $0.8896 \pm 0.0957$
& $0.0879 \pm 0.0574$
& $0.1521 \pm 0.0573$ \\
& Habitats
& $\mathbf{0.9140 \pm 0.0113}$
& $0.9442 \pm 0.0690$
& $\mathbf{0.1585 \pm 0.0698}$
& $\mathbf{0.2186 \pm 0.0602}$ \\
\midrule

\multirow{3}{*}{Macro-average}
& Independent Generation
& $0.7400$
& $0.7471$
& $0.1899$
& $0.1552$ \\
& Direct Mutation
& $\mathbf{0.8409}$
& $0.8222$
& $0.2146$
& $0.2180$ \\
& Habitats
& $0.7787$
& $\mathbf{0.8533}$
& $\mathbf{0.3355}$
& $\mathbf{0.3825}$ \\
\bottomrule
\end{tabular}
\end{table*}

\begin{table*}[h]
\centering
\caption{
Complete Grounded Behavioral Profile Coverage (GC) results.
}
\label{tab:full_gc_results}
\small
\setlength{\tabcolsep}{8pt}
\begin{tabular}{llcc}
\toprule
\textbf{Environment}
& \textbf{Method}
& $\boldsymbol{\mathrm{GC}_{\mathrm{AUC}}}$
& \textbf{Final GC} \\
\midrule
\multirow{3}{*}{Card Sandbox}
& Independent Generation
& $211.33 \pm 36.56$
& $23.67 \pm 5.31$ \\
& Direct Mutation
& $244.33 \pm 38.64$
& $23.33 \pm 6.60$ \\
& Habitats
& $\mathbf{322.67} \pm 37.89$
& $\mathbf{26.00} \pm 2.83$ \\
\midrule
\multirow{3}{*}{Cart-Pole}
& Independent Generation
& $7 \pm 5$
& $1 \pm 1$ \\
& Direct Mutation
& $\mathbf{16} \pm 10$
& $\mathbf{2} \pm 1$ \\
& Habitats
& $6 \pm 6$
& $\mathbf{2} \pm 2$ \\
\midrule
\multirow{3}{*}{Hopper}
& Independent Generation
& $7 \pm 2$
& $1 \pm 1$ \\
& Direct Mutation
& $2 \pm 4$
& $0 \pm 1$ \\
& Habitats
& $\mathbf{16} \pm 14$
& $\mathbf{3} \pm 3$ \\
\midrule
\multirow{3}{*}{Cheetah}
& Independent Generation
& $18 \pm 1$
& $4 \pm 2$ \\
& Direct Mutation
& $21 \pm 7$
& $4 \pm 2$ \\
& Habitats
& $\mathbf{29} \pm 3$
& $\mathbf{7} \pm 2$ \\
\bottomrule
\end{tabular}
\end{table*}

%%%%%%%%%%%%%%%%%%%%%%%%%%%%%%%%%%%%%%%%%%%%%%%%%%%%%%%%%%%%%%%%
% New Appendix
%%%%%%%%%%%%%%%%%%%%%%%%%%%%%%%%%%%%%%%%%%%%%%%%%%%%%%%%%%%%%%%%
\newpage
\section{Downstream Adaptation Evaluation}
\label{app:meta_evaluation}

\subsection{Downstream Tasks}
\label{app:downstream_tasks}

We evaluate whether evolved policy libraries support downstream learning in the
Card Sandbox. The ten downstream task specifications were fixed before
evolution and were never explicitly provided through the evolution prompts,
policy-training objectives, or survivor-selection procedure. All tasks use the
same environment dynamics and primitive action interface while defining
distinct downstream reward functions.

\begin{table}[h]
\centering
\caption{
Downstream Card Sandbox tasks used to evaluate adaptation with evolved policy
libraries.
}
\label{tab:downstream_tasks}
\small
\setlength{\tabcolsep}{4pt}
\begin{tabular}{p{0.31\linewidth}p{0.61\linewidth}}
\toprule
\textbf{Task} & \textbf{Objective} \\
\midrule
Ascending Sort
& Play cards in non-decreasing order. \\

Descending Sort
& Play cards in non-increasing order. \\

Triplet Matching
& Construct groups of three cards with matching values. \\

Target Sum
& Select cards whose cumulative value reaches a specified target. \\

High-Card Maximization
& Maximize the cumulative value of the cards played. \\

Low-Card Minimization
& Preferentially play cards with the smallest available values. \\

Even--Odd Pattern
& Produce sequences that alternate between even- and odd-valued cards. \\

Sandwich Pair
& Construct patterns of the form \(A\text{--}B\text{--}A\). \\

Rapid Clear Discard
& Empty the hand as quickly as possible. \\

Full-House Triple--Pair
& Construct a combination containing one matching triplet and one matching
pair. \\
\bottomrule
\end{tabular}
\end{table}

\subsection{Policy Libraries and Downstream Agents}
\label{app:downstream_agents}

For each evolutionary method and seed, we construct a fixed library by
uniformly sampling 15 policies without replacement from the final survivor
population. This size-matches the Habitats, Direct Mutation, and Independent
Generation libraries. Selection is performed without reference to downstream
performance. These policies remain frozen throughout downstream training. A
hierarchical meta-policy observes the environment state and selects one library
policy at each step; the selected policy then produces the primitive
environment action.

We compare this architecture with a tabular Q-learning policy, which we call
Scratch, trained directly on each downstream reward using the same learning
algorithm and optimization hyperparameters as the Card Sandbox policies.
Meta-policies over each evolved library are trained using downstream budgets
$B_{\mathrm{Meta}}\in\{1{,}000,2{,}000,5{,}000,10{,}000,20{,}000\}$ episodes,
while Scratch is trained for $B_{\mathrm{Scratch}}=20{,}000$ episodes. All
library conditions use the same meta-policy architecture, exploration-schedule
construction, evaluation protocol, and matched random seeds.

This comparison measures adaptation efficiency conditional on an already
evolved policy library; it does not compare total computation, including the
cost of evolving that library.

\subsection{Evaluation Protocol and Win Rate}
\label{app:downstream_protocol}

We evaluate meta-policies over the Habitats, Direct Mutation, and Independent
Generation libraries, together with the matched Scratch agent, on all ten
downstream tasks using three matched seeds. For each method and seed, a
separate meta-policy is trained at each downstream budget and compared with the
Scratch policy trained for $20{,}000$ episodes. Per-task performance is
measured by mean return over 50 evaluation episodes.

Because reward scales differ across tasks, we aggregate performance using
paired win rate rather than averaging raw returns. For library source
$M\in\{\mathrm{Habitats},\mathrm{Direct},\mathrm{Independent}\}$, evolution
seed $s$, and training budget
$B\in\{1{,}000,2{,}000,5{,}000,10{,}000,20{,}000\}$, let $W_{M,s,B}$,
$T_{M,s,B}$, and $L_{M,s,B}$ denote the numbers of tasks on which the
corresponding meta-policy respectively outperforms, ties, or underperforms the
matched Scratch policy trained for $20{,}000$ episodes. A tie is recorded when
their paired mean returns are equal. We define
\begin{equation}
\mathrm{WR}_{M,s,B}
=
\frac{
W_{M,s,B} + \tfrac{1}{2}T_{M,s,B}
}{
W_{M,s,B} + T_{M,s,B} + L_{M,s,B}
}.
\label{eq:downstream_win_rate}
\end{equation}
Thus, $\mathrm{WR}=0.5$ represents parity with Scratch. We compute win rate
independently for each method, seed, and budget and report its mean and
standard deviation across the three seeds.

\subsection{Complete Downstream Results}
\label{app:downstream_results}

Tables~\ref{tab:downstream_adaptation_results}--\ref{tab:downstream_independent_results}
report complete per-task returns for meta-policies trained over each evolved
library. Because reward scales and success criteria differ across tasks, raw
returns should be compared only within each row. At $5{,}000$ episodes,
Habitats and Direct Mutation approach parity with Scratch, achieving paired
win rates of $0.467 \pm 0.029$ and $0.483 \pm 0.029$, respectively. At
$10{,}000$ episodes, Habitats reaches $0.617 \pm 0.058$, compared with
$0.533 \pm 0.029$ for Direct Mutation and $0.467 \pm 0.058$ for Independent
Generation. Habitats at $10{,}000$ episodes also exceeds both baseline
libraries after $20{,}000$ episodes, demonstrating more sample-efficient
downstream reuse.

\begin{table*}[h]
\centering
\caption{
Complete downstream-adaptation results for the Habitats library. Meta-policy
columns vary the downstream training budget, while Scratch is trained for
\(20{,}000\) episodes. Entries report mean return \(\pm\) standard deviation
across three matched seeds. Within each task, the highest mean is shown in
bold. The final row reports paired win rate against Scratch, computed
independently within each seed using Eq.~\eqref{eq:downstream_win_rate}.
}
\label{tab:downstream_adaptation_results}
\small
\setlength{\tabcolsep}{5pt}
\resizebox{\textwidth}{!}{%
\begin{tabular}{lcccccc}
\toprule
& \multicolumn{5}{c}{\textbf{Habitats meta-policy}}
& \textbf{Scratch} \\
\cmidrule(lr){2-6}
\cmidrule(lr){7-7}
\textbf{Downstream task}
& \textbf{1k}
& \textbf{2k}
& \textbf{5k}
& \textbf{10k}
& \textbf{20k}
& \textbf{20k} \\
\midrule
Ascending Sort
& \(3.27 \pm 1.30\)
& \(8.67 \pm 0.23\)
& \(9.13 \pm 1.50\)
& \(\boldsymbol{10.00 \pm 0.00}\)
& \(\boldsymbol{10.00 \pm 0.00}\)
& \(\boldsymbol{10.00 \pm 0.00}\) \\

Descending Sort
& \(-2.13 \pm 4.23\)
& \(5.35 \pm 4.33\)
& \(9.73 \pm 0.46\)
& \(\boldsymbol{10.00 \pm 0.00}\)
& \(\boldsymbol{10.00 \pm 0.00}\)
& \(\boldsymbol{10.00 \pm 0.00}\) \\

Triplet Matching
& \(0.00 \pm 0.00\)
& \(1.07 \pm 1.85\)
& \(6.13 \pm 2.89\)
& \(\boldsymbol{10.00 \pm 0.00}\)
& \(\boldsymbol{10.00 \pm 0.00}\)
& \(\boldsymbol{10.00 \pm 0.00}\) \\

Target Sum
& \(0.40 \pm 0.69\)
& \(4.33 \pm 1.10\)
& \(9.20 \pm 0.72\)
& \(\boldsymbol{10.00 \pm 0.00}\)
& \(\boldsymbol{10.00 \pm 0.00}\)
& \(8.36 \pm 0.00\) \\

High-Card Maximization
& \(3.49 \pm 3.73\)
& \(27.35 \pm 0.07\)
& \(\boldsymbol{31.12 \pm 0.05}\)
& \(\boldsymbol{31.12 \pm 0.05}\)
& \(\boldsymbol{31.12 \pm 0.05}\)
& \(30.70 \pm 0.00\) \\

Low-Card Minimization
& \(0.80 \pm 0.69\)
& \(6.53 \pm 1.45\)
& \(\boldsymbol{10.00 \pm 0.00}\)
& \(\boldsymbol{10.00 \pm 0.00}\)
& \(\boldsymbol{10.00 \pm 0.00}\)
& \(\boldsymbol{10.00 \pm 0.00}\) \\

Even--Odd Pattern
& \(2.20 \pm 2.20\)
& \(7.87 \pm 1.94\)
& \(\boldsymbol{10.00 \pm 0.00}\)
& \(\boldsymbol{10.00 \pm 0.00}\)
& \(\boldsymbol{10.00 \pm 0.00}\)
& \(\boldsymbol{10.00 \pm 0.00}\) \\

Sandwich Pair
& \(0.87 \pm 0.42\)
& \(3.73 \pm 2.25\)
& \(9.93 \pm 0.12\)
& \(\boldsymbol{10.00 \pm 0.00}\)
& \(\boldsymbol{10.00 \pm 0.00}\)
& \(6.72 \pm 0.00\) \\

Rapid Clear Discard
& \(-3.43 \pm 1.36\)
& \(-2.40 \pm 2.28\)
& \(0.27 \pm 1.06\)
& \(6.33 \pm 0.82\)
& \(6.41 \pm 0.68\)
& \(\boldsymbol{6.80 \pm 0.00}\) \\

Full-House Triple--Pair
& \(0.00 \pm 0.00\)
& \(0.53 \pm 0.61\)
& \(4.56 \pm 3.63\)
& \(7.76 \pm 1.18\)
& \(7.96 \pm 1.04\)
& \(\boldsymbol{10.00 \pm 0.00}\) \\
\midrule
\textbf{Win rate vs.\ Scratch 20k}
& \(0.0\% \pm 0.0\%\)
& \(3.3\% \pm 2.9\%\)
& \(46.7\% \pm 2.9\%\)
& \(\boldsymbol{61.7\% \pm 5.8\%}\)
& \(\boldsymbol{61.7\% \pm 5.8\%}\)
& -- \\
\bottomrule
\end{tabular}%
}
\end{table*}

\begin{table*}[h]
\centering
\caption{
Complete downstream-adaptation results for the Direct Mutation library.
Meta-policy columns vary the downstream training budget, while Scratch is
trained for \(20{,}000\) episodes. Entries report mean return \(\pm\) standard
deviation across three matched seeds. Within each task, the highest mean is
shown in bold. The final row reports paired win rate against Scratch.
}
\label{tab:downstream_direct_results}
\small
\setlength{\tabcolsep}{5pt}
\resizebox{\textwidth}{!}{%
\begin{tabular}{lcccccc}
\toprule
& \multicolumn{5}{c}{\textbf{Direct Mutation meta-policy}}
& \textbf{Scratch} \\
\cmidrule(lr){2-6}
\cmidrule(lr){7-7}
\textbf{Downstream task}
& \textbf{1k}
& \textbf{2k}
& \textbf{5k}
& \textbf{10k}
& \textbf{20k}
& \textbf{20k} \\
\midrule
Ascending Sort
& \(0.67 \pm 2.31\)
& \(3.20 \pm 2.11\)
& \(\boldsymbol{10.00 \pm 0.00}\)
& \(\boldsymbol{10.00 \pm 0.00}\)
& \(\boldsymbol{10.00 \pm 0.00}\)
& \(\boldsymbol{10.00 \pm 0.00}\) \\

Descending Sort
& \(0.87 \pm 0.90\)
& \(2.07 \pm 3.58\)
& \(\boldsymbol{10.00 \pm 0.00}\)
& \(\boldsymbol{10.00 \pm 0.00}\)
& \(\boldsymbol{10.00 \pm 0.00}\)
& \(\boldsymbol{10.00 \pm 0.00}\) \\

Triplet Matching
& \(-2.53 \pm 4.39\)
& \(0.00 \pm 0.00\)
& \(3.33 \pm 1.42\)
& \(9.61 \pm 0.67\)
& \(9.61 \pm 0.67\)
& \(\boldsymbol{10.00 \pm 0.00}\) \\

Target Sum
& \(0.33 \pm 0.31\)
& \(1.07 \pm 1.85\)
& \(9.20 \pm 0.69\)
& \(\boldsymbol{10.00 \pm 0.00}\)
& \(\boldsymbol{10.00 \pm 0.00}\)
& \(8.36 \pm 0.00\) \\

High-Card Maximization
& \(0.21 \pm 0.36\)
& \(12.45 \pm 7.63\)
& \(\boldsymbol{31.15 \pm 0.05}\)
& \(\boldsymbol{31.15 \pm 0.05}\)
& \(\boldsymbol{31.15 \pm 0.05}\)
& \(30.70 \pm 0.00\) \\

Low-Card Minimization
& \(0.00 \pm 0.00\)
& \(1.07 \pm 1.85\)
& \(8.73 \pm 2.19\)
& \(\boldsymbol{10.00 \pm 0.00}\)
& \(\boldsymbol{10.00 \pm 0.00}\)
& \(\boldsymbol{10.00 \pm 0.00}\) \\

Even--Odd Pattern
& \(1.07 \pm 0.95\)
& \(1.64 \pm 2.77\)
& \(\boldsymbol{10.00 \pm 0.00}\)
& \(\boldsymbol{10.00 \pm 0.00}\)
& \(\boldsymbol{10.00 \pm 0.00}\)
& \(\boldsymbol{10.00 \pm 0.00}\) \\

Sandwich Pair
& \(0.40 \pm 0.53\)
& \(0.07 \pm 0.12\)
& \(9.33 \pm 0.61\)
& \(\boldsymbol{10.00 \pm 0.00}\)
& \(\boldsymbol{10.00 \pm 0.00}\)
& \(6.72 \pm 0.00\) \\

Rapid Clear Discard
& \(-13.70 \pm 8.36\)
& \(-6.04 \pm 6.55\)
& \(2.25 \pm 1.94\)
& \(6.55 \pm 0.24\)
& \(6.78 \pm 0.02\)
& \(\boldsymbol{6.80 \pm 0.00}\) \\

Full-House Triple--Pair
& \(0.00 \pm 0.00\)
& \(0.00 \pm 0.00\)
& \(5.79 \pm 1.09\)
& \(7.57 \pm 1.50\)
& \(8.36 \pm 0.16\)
& \(\boldsymbol{10.00 \pm 0.00}\) \\
\midrule
\textbf{Win rate vs.\ Scratch 20k}
& \(0.0\% \pm 0.0\%\)
& \(0.0\% \pm 0.0\%\)
& \(48.3\% \pm 2.9\%\)
& \(53.3\% \pm 2.9\%\)
& \(\boldsymbol{56.7\% \pm 2.9\%}\)
& -- \\
\bottomrule
\end{tabular}%
}
\end{table*}

\begin{table*}[h]
\centering
\caption{
Complete downstream-adaptation results for the Independent Generation
library. Meta-policy columns vary the downstream training budget, while
Scratch is trained for \(20{,}000\) episodes. Entries report mean return
\(\pm\) standard deviation across three matched seeds. Within each task, the
highest mean is shown in bold. The final row reports paired win rate against
Scratch.
}
\label{tab:downstream_independent_results}
\small
\setlength{\tabcolsep}{5pt}
\resizebox{\textwidth}{!}{%
\begin{tabular}{lcccccc}
\toprule
& \multicolumn{5}{c}{\textbf{Independent Generation meta-policy}}
& \textbf{Scratch} \\
\cmidrule(lr){2-6}
\cmidrule(lr){7-7}
\textbf{Downstream task}
& \textbf{1k}
& \textbf{2k}
& \textbf{5k}
& \textbf{10k}
& \textbf{20k}
& \textbf{20k} \\
\midrule
Ascending Sort
& \(3.61 \pm 4.86\)
& \(7.07 \pm 5.08\)
& \(\boldsymbol{10.00 \pm 0.00}\)
& \(\boldsymbol{10.00 \pm 0.00}\)
& \(\boldsymbol{10.00 \pm 0.00}\)
& \(\boldsymbol{10.00 \pm 0.00}\) \\

Descending Sort
& \(4.00 \pm 4.06\)
& \(6.00 \pm 5.10\)
& \(\boldsymbol{10.00 \pm 0.00}\)
& \(\boldsymbol{10.00 \pm 0.00}\)
& \(\boldsymbol{10.00 \pm 0.00}\)
& \(\boldsymbol{10.00 \pm 0.00}\) \\

Triplet Matching
& \(-1.24 \pm 2.15\)
& \(-1.24 \pm 2.15\)
& \(0.13 \pm 0.23\)
& \(0.27 \pm 0.23\)
& \(0.20 \pm 0.35\)
& \(\boldsymbol{10.00 \pm 0.00}\) \\

Target Sum
& \(2.60 \pm 6.01\)
& \(4.13 \pm 3.13\)
& \(8.80 \pm 2.08\)
& \(\boldsymbol{9.93 \pm 0.12}\)
& \(\boldsymbol{9.93 \pm 0.12}\)
& \(8.36 \pm 0.00\) \\

High-Card Maximization
& \(4.89 \pm 8.46\)
& \(15.35 \pm 15.25\)
& \(\boldsymbol{30.97 \pm 0.42}\)
& \(\boldsymbol{30.97 \pm 0.42}\)
& \(\boldsymbol{30.97 \pm 0.42}\)
& \(30.70 \pm 0.00\) \\

Low-Card Minimization
& \(4.47 \pm 5.08\)
& \(5.27 \pm 5.02\)
& \(9.08 \pm 1.59\)
& \(\boldsymbol{10.00 \pm 0.00}\)
& \(\boldsymbol{10.00 \pm 0.00}\)
& \(\boldsymbol{10.00 \pm 0.00}\) \\

Even--Odd Pattern
& \(5.40 \pm 4.81\)
& \(6.73 \pm 4.66\)
& \(9.80 \pm 0.35\)
& \(9.87 \pm 0.23\)
& \(9.87 \pm 0.23\)
& \(\boldsymbol{10.00 \pm 0.00}\) \\

Sandwich Pair
& \(-0.45 \pm 0.96\)
& \(1.67 \pm 2.89\)
& \(8.43 \pm 1.67\)
& \(\boldsymbol{9.69 \pm 0.53}\)
& \(9.64 \pm 0.62\)
& \(6.72 \pm 0.00\) \\

Rapid Clear Discard
& \(-15.57 \pm 8.90\)
& \(-9.60 \pm 4.46\)
& \(2.17 \pm 3.22\)
& \(6.23 \pm 0.96\)
& \(6.07 \pm 1.23\)
& \(\boldsymbol{6.80 \pm 0.00}\) \\

Full-House Triple--Pair
& \(2.09 \pm 3.63\)
& \(1.17 \pm 2.03\)
& \(4.56 \pm 2.40\)
& \(8.49 \pm 0.09\)
& \(8.49 \pm 0.09\)
& \(\boldsymbol{10.00 \pm 0.00}\) \\
\midrule
\textbf{Win rate vs.\ Scratch 20k}
& \(6.7\% \pm 11.5\%\)
& \(6.7\% \pm 7.6\%\)
& \(36.7\% \pm 15.3\%\)
& \(\boldsymbol{46.7\% \pm 5.8\%}\)
& \(\boldsymbol{46.7\% \pm 5.8\%}\)
& -- \\
\bottomrule
\end{tabular}%
}
\end{table*}

%%%%%%%%%%%%%%%%%%%%%%%%%%%%%%%%%%%%%%%%%%%%%%%%%%%%%%%%%%%%%%%%
% Qualitative Examples
%%%%%%%%%%%%%%%%%%%%%%%%%%%%%%%%%%%%%%%%%%%%%%%%%%%%%%%%%%%%%%%%
\clearpage
\section{Qualitative Examples}
\label{app:qualitative_examples}

The aggregate metrics characterize whether Habitats produces high-quality,
behaviorally distinct, and grounded populations, but they do not convey what
the discovered objectives actually require. We therefore inspect representative
reward programs, induced policy specializations, and niche-replacement events
from the evolutionary traces. These examples were selected from the evaluated
runs and were not generated retrospectively.

\subsection{Discrete Card Sandbox}
\label{app:qualitative_card}

We inspect representative reward programs and induced behaviors from three
independent Habitats runs in the Card Sandbox. Let $(x_0,x_1,x_2)$ denote the
values occupying the three table positions and let $h$ denote the card
currently held by the agent. The conditions below summarize the principal
objective implemented by each reward program; the executable rewards may also
contain intermediate shaping terms that facilitate learning.

\paragraph{Breadth of discovered objectives.}
Despite operating within the same compact environment, the generated rewards
elicited markedly different ways of manipulating the cards. Some objectives
were based on ordering relations. For example, \emph{The Sorting Table}
rewarded the strictly increasing arrangement $x_0<x_1<x_2$, while
\emph{The Peak Pattern} instead required the non-monotonic configuration
$x_0<x_1>x_2$. Other objectives emphasized arithmetic structure.
\emph{The Summation Table} required $x_0+x_1=x_2$, whereas
\emph{The Table Product} coupled the held card to the table through
$h=x_0x_1$. \emph{The Magic Fifteen} introduced a global constraint over the
complete active configuration, $h+x_0+x_1+x_2=15$.

Habitats also discovered objectives based on repetition and symmetry.
\emph{The Perfect Triplet} required three cards of the same value, while
\emph{The Symmetric Sandwich} required matching outer cards surrounding a
different center, $x_0=x_2\neq x_1$. Although both objectives use the same
cards and primitive actions, they demand different forms of coordination: one
policy must collect repeated values, whereas the other must preserve both
equality and position.

Other rewards relied on number-theoretic properties absent from the simpler
ordering and matching objectives. \emph{Distinct Even Trio} required three
pairwise-distinct even cards, while \emph{Strict Divisibility Descent} required
a decreasing sequence satisfying $x_0\bmod x_1=0$ and
$x_1\bmod x_2=0$. The discovered objectives therefore extended beyond
variations of sorting and matching to include parity, primality, arithmetic
progressions, target sums, products, and modular relations.

\paragraph{Increasing compositionality.}
Across the evolutionary traces, later objectives frequently combined
relations that appeared separately in simpler rewards. The progression from
\emph{The Sorting Table} to \emph{The Stepping Stones} illustrates this
pattern. Whereas the former requires only ascending order, the latter also
requires equal spacing, $x_1-x_0=x_2-x_1$. \emph{The Consecutive Run} further
specializes this structure by fixing the spacing to one:
$x_1=x_0+1$ and $x_2=x_1+1$.

A similar elaboration occurred among sandwich objectives. Starting from the
basic symmetry of \emph{The Symmetric Sandwich}, Habitats generated
\emph{The High--Low Sandwich}, which restricts the outer cards to high values
and the center to a low value. Other variants imposed parity or number-class
constraints. \emph{The High-Odd/Low-Even Sandwich}, for example, requires
high odd cards in the outer positions and a low even card in the center,
whereas \emph{The Composite Sandwich} restricts the pattern to composite
values. The same compositional pattern appeared in
\emph{The Prime Peak Pattern}, which combines the non-monotonic structure of
\emph{The Peak Pattern} with the requirement that every participating card be
prime.

These examples show that mutation did more than rename existing objectives or
adjust their numerical coefficients. It introduced new relational structure
and composed previously useful concepts into increasingly constrained
mini-games. The resulting rewards require policies to coordinate several
properties simultaneously, including card value, position, parity, and
arithmetic relation.

\paragraph{Cross-reward generalization.}
Cross-evaluation reveals when a policy trained under one generated reward
acquires behavior that is also useful under another. Because selection occurs
within reward-defined niches, a newly generated individual can replace an
incumbent specialist by performing better on the incumbent's own objective.
These replacement events provide qualitative evidence that the induced
behaviors are not confined to the rewards under which they were learned.

The clearest example appeared in the first run, where a policy trained on
\emph{The Composite Arithmetic Solver} displaced specialists associated with
\emph{The Sorting Table}, \emph{The Stepping Stones},
\emph{The Divisibility Pair}, and \emph{The Consecutive Run}. The challenger
had been trained on an objective combining factor-product and
arithmetic-progression structure, yet its resulting behavior transferred
across several simpler ordering and arithmetic niches.

This transfer does not imply that the corresponding reward programs are
logically equivalent or that one constraint formally contains another.
Selection operates on realized policy behavior rather than textual or symbolic
similarity between reward functions. Habitats can therefore expose functional
relationships that would not necessarily be apparent from the reward names or
source code alone.

Comparable niche replacements appeared in the other runs. In the second run,
\emph{The Composite Solver} replaced the \emph{Peak Pattern} specialist. In
the third, \emph{Strict Divisibility Descent} displaced
\emph{Ascending Prime Straight}, while \emph{Distinct Even Trio} later
displaced \emph{Parity Alternator}. The continued appearance of such events
suggests that cross-niche competition remained active as new objectives and
behaviors entered the population.

\paragraph{Takeover and re-diversification.}
Broadly capable policies can replace several incumbent specialists within a
short portion of an evolutionary trace, temporarily narrowing the range of
behaviors represented in the surviving population. The first run provides the
clearest example. The emergence of \emph{The Composite Arithmetic Solver}
produced a concentrated takeover across multiple existing niches, leaving a
smaller collection of broadly successful behaviors.

This contraction did not persist. Subsequent generations introduced objectives
including \emph{Descending Even Run}, \emph{Prime Sandwich},
\emph{Composite Sandwich}, \emph{Prime Peak Pattern},
\emph{High-Odd/Low-Even Sandwich}, \emph{Strict Divisibility Descent}, and
\emph{Distinct Even Trio}. These objectives introduced new combinations of
direction, magnitude, parity, primality, and divisibility, restoring a broad
range of policy specializations after the generalist takeover. The trace
therefore exhibits a qualitative cycle of specialization, generalist
replacement, population contraction, and renewed diversification.

The second run displayed a similar but later turnover. Its population first
accumulated objectives such as \emph{Addition Equalizer},
\emph{Ascending Odd Straight}, \emph{Parity Sandwich}, and
\emph{Prime Sum}. A subsequent takeover replaced several established
specialists: \emph{The Composite Solver} displaced
\emph{The Peak Pattern}, while policies trained on
\emph{Parity Sandwich} and \emph{Descending Divisibility Chain} both
displaced the \emph{Perfect Triplet} specialist. Later offspring again
introduced distinct structural constraints, preventing the takeover from
terminating objective discovery altogether.

\paragraph{Stable diversification.}
The third run followed a different qualitative trajectory. Rather than
undergoing a pronounced generalist takeover, it retained a broad mixture of
ordering, matching, prime-constrained, parity-based, and divisibility-based
objectives throughout evolution. The population nevertheless remained
dynamic. \emph{The Consecutive Run} displaced the
\emph{Ascending Straight} specialist, while policies associated with
\emph{Three of a Kind}, \emph{Ascending Prime Straight}, and
\emph{Odd Sandwich} each transferred successfully to the
\emph{Prime Sum} niche. Later replacements involving
\emph{Strict Divisibility Descent} and \emph{Distinct Even Trio} show that
continued objective discovery and competitive turnover can also occur without
a sharp population bottleneck.

\paragraph{Summary.}
Taken together, these examples reveal an evolving collection of structurally
different objectives rather than variations of a single reward template.
Later rewards frequently combine relations introduced separately by earlier
objectives, producing increasingly compositional policy specializations.
Cross-evaluation also creates meaningful population-level competition: new
policies can replace incumbent specialists on the incumbents' own rewards,
sometimes producing temporary takeovers followed by the emergence of new
niches. The resulting population is therefore better understood as an
evolving ecology of objectives and behaviors than as a collection of
independently generated reward programs.

\subsection{Continuous Cart-Pole}
Cart-Pole generated multiple different individuals that learned to master their own niches.

\paragraph{Breadth of discovered objectives.}

\emph{The Smooth Omnidirectional Spinner} required the agent to maintain rapid, continuous pole rotation in either the clockwise or counter-clockwise direction. The reward program incentivized the absolute value of the pole's angular velocity, making the direction of spin irrelevant and rewarding only sustained rotational speed. Rather than imposing a hard positional constraint or early termination when the cart drifted, the reward employed a smooth Gaussian-style centering term that provided a stable, non-zero gradient across the entire track length.

\emph{The Quiet Centered Balancer}, derived from \emph{The Smooth Omnidirectional
Spinner}, required the agent to maintain a motionless, vertically upright pole at the center of the track. The reward maximized the cosine of the pole angle (peaking when the pole points straight up) while uniformly penalizing angular velocity to suppress
oscillations. It inherited the smooth Gaussian centering term from \emph{The Smooth Omnidirectional Spinner}, but replaced the spin incentive with a stillness objective. The resulting policy swung the pole upward through smooth, controlled adjustments and then held it in a quiet equilibrium, producing a fundamentally different motor strategy from
the continuous-rotation policy despite sharing the same centering mechanism.

\emph{The Downward Stroller} required the agent to keep the pole hanging straight down (the minimum-energy equilibrium) while maintaining a steady cruising speed along the track. The reward combined a downward-orientation measure that peaked when the pole pointed vertically downward, a Gaussian velocity reward centered on a moderate target speed, and a soft boundary penalty that increased smoothly as the cart approached the track edges. This individual represents an independent line of discovery that inverts the conventional understanding of the environment. Standard Cart-Pole objectives treat the downward-hanging configuration as a failure state; this reward treats it as the goal. By coupling the inverted orientation target with a constant-velocity objective and soft boundary awareness, the reward constructs a mini-task that exercises control capabilities (smooth velocity regulation, gentle reversal, boundary avoidance) entirely absent from both the spinning and balancing policies.

\paragraph{Increasing compositionality.}
The lineage from \emph{The Smooth Omnidirectional Spinner} to \emph{The
Quiet Centered Balancer} provides a clear example of compositional
inheritance within the evolutionary trace. The Spinner introduced two
innovations relative to earlier reward programs: direction-agnostic rotation
reward via the absolute angular velocity, and smooth Gaussian centering in
place of hard positional constraints. The Balancer retained the centering
mechanism but replaced the rotation objective with an orientation-and-stillness
objective. This selective recombination mirrors the compositional
elaboration observed in the Card Sandbox, where later rewards combined
relations introduced separately by earlier objectives.

\paragraph{Physical contrast and skill grounding.}
The three objectives partition the environment's state space into
non-overlapping behavioral regimes.  The Spinner maximizes angular velocity
and is indifferent to pole orientation; the Balancer minimizes angular
velocity and targets the upward equilibrium; the Stroller minimizes angular
velocity but targets the downward equilibrium while simultaneously regulating
translational speed. Each policy must coordinate cart motion with pole
dynamics, but the nature of that coordination differs: the Spinner uses the
cart to sustain rotational momentum, the Balancer uses it to damp
perturbations around an unstable fixed point, and the Stroller uses it to
cruise at constant speed without disturbing a stable fixed point.  These
three strategies collectively demonstrate that Habitats can discover
objectives spanning different dynamical regimes from a single environment.

\subsection{Continuous Cheetah}
\label{app:qualitative_cheetah}

\paragraph{Breadth of discovered objectives.}
The generated rewards induced locomotion modes that differ not only in speed
but in the fundamental physical strategy employed by the agent.
\emph{The Commando Crawl} required the agent to slide and drag itself forward
as rapidly as possible while keeping its torso strictly below a low elevation
threshold and maintaining a perfectly flat body orientation.  The reward
program achieved this through a combination of forward-speed incentive,
a steep quadratic penalty on torso elevation above the threshold, and an
orientation penalty that suppressed tumbling or flipping during the crawl.
A small action-magnitude regularizer discouraged high-frequency joint
vibrations, yielding a smooth, belly-down dragging gait that is qualitatively
unlike any standard running policy.

In contrast, \emph{The Trampoline Jumper} required the agent to launch itself
vertically as high as possible while remaining perfectly stationary in the
horizontal direction and maintaining a stable, upright posture.  This reward
combined an upward-velocity incentive with a bonus for achieving elevation
above the baseline standing height, a heavy penalty on any horizontal
displacement, orientation and angular-velocity penalties to keep the torso
flat during flight, and a control-effort regularizer for smooth coordination.
The resulting policy produced repeated vertical hops with negligible
lateral drift, a locomotion mode that shares no kinematic features with the
crawling specialist.

\paragraph{Locomotion relevance and skill grounding.}
These two objectives illustrate how Habitats can decompose the latent
capability space of a single morphology into complementary skills. Crawling and jumping are both physically meaningful locomotion primitives: one minimizes the body's vertical profile while maximizing ground contact, and the other maximizes vertical impulse while suppressing horizontal motion. Neither objective was specified by the experimenter; both emerged from the evolutionary process operating over the same action space and dynamics.

The diversity of the discovered rewards is particularly notable because the
Cheetah morphology is often regarded as a ``running'' environment. Habitats demonstrates that the body can be repurposed for fundamentally different types of movement when the objective function is varied.

\subsection{Continuous Hopper}
\label{app:qualitative_hopper}

\paragraph{Breadth of discovered objectives.}
\emph{The Gazelle} required the agent to execute explosive, long-distance
forward bounds by coupling forward speed with vertical elevation. The reward
program terminated episodes when the torso dropped below a minimum height,
preventing dragging or collapse. Through a combination of uprightness
multipliers, decay factors, and vertical bonuses, the policy learned to convert
vertical thrust into long parabolic arcs rather than low, shuffling hops.

\emph{The Giant Strider} instead targeted high forward velocity while imposing
a substantial penalty on joint velocities. This discouraged rapid,
high-frequency actuation and forced the policy to produce speed through long,
controlled strides. The resulting behavior covered ground efficiently through
broad, sweeping motions rather than the explosive aerial bounds learned by
\emph{The Gazelle}.

\emph{The Pogo Stick} required the agent to hop as high as possible while
remaining stationary and upright. The reward applied a quadratic height bonus
above a standing threshold, making dynamic jumping more profitable than merely
standing tall. This height reward was gated by an uprightness factor that
vanished when the torso tilted beyond a moderate angular threshold, while a
horizontal-speed penalty suppressed lateral drift.

\paragraph{Contrasting physical strategies.}
The three objectives divide the Hopper's dynamics along complementary axes.
\emph{The Gazelle} combines vertical thrust with forward movement to produce
explosive bounds. \emph{The Giant Strider} pursues forward speed under a
joint-velocity constraint, favoring long and controlled ground-covering
motions. \emph{The Pogo Stick} treats horizontal displacement as a liability
and redirects the same morphology's thrust capacity into repeated vertical
launches.

These behaviors demonstrate that Habitats can discover several physically
distinct strategies from the same embodiment: aerial bounding, efficient
striding, and stationary vertical hopping. Their coexistence within one
evolutionary population shows that objective evolution can separate locomotion into complementary behavioral primitives without an experimenter-specified skill taxonomy or curriculum.

%%%%%%%%%%%%%%%%%%%%%%%%%%%%%%%%%%%%%%%%%%%%%%%%%%%%%%%%%%%%%%%%
% Evolutionary-Horizon Analysis
%%%%%%%%%%%%%%%%%%%%%%%%%%%%%%%%%%%%%%%%%%%%%%%%%%%%%%%%%%%%%%%%
\clearpage
\section{Evolutionary-Horizon Analysis}
\label{app:horizon_analysis}

To test whether the main results depend on the chosen evolutionary horizons,
we extend Card Sandbox from $15$ to $30$ generations and Cart-Pole, Hopper,
and Cheetah from $6$ to $10$ generations. In order to make sure continuous individuals have enough time to learn their reward, we use the same training configurations as those given by the Brax engine \cite{freeman2021braxdifferentiablephysics} for each specific MuJoCo environments. Those configurations are given in Table \ref{tab:mujoco_hyperparams}.

\begin{table*}[h]
\centering
\caption{Policy-training and evaluation hyperparameters for MuJoCo continuous control environments.}
\label{tab:mujoco_hyperparams}
\small
\setlength{\tabcolsep}{5pt}
\begin{tabular}{lccc}
\toprule
\textbf{Hyperparameter}
& \textbf{Cartpole}
& \textbf{Cheetah}
& \textbf{Hopper}\\
\midrule
Policy optimizer
& \multicolumn{3}{c}{PPO (Brax / JAX)}\\

Learning rate
& $3\times10^{-4}$
& $3\times10^{-4}$
& $6\times10^{-4}$\\

Discount factor $\gamma$
& $0.97$
& $0.95$
& $0.997$\\

Exploration or clipping
& \multicolumn{3}{c}{PPO clip ratio $\epsilon = 0.2$}\\

Training budget per new policy
& $10{,}000{,}000$ steps
& $50{,}000{,}000$ steps
& $50{,}000{,}000$ steps\\

Evaluation rollouts $E_{\mathrm{eval}}$
& $20$ episodes
& $20$ episodes
& $20$ episodes\\

Train episode horizon $H$
& $1{,}000$ steps
& $1{,}000$ steps
& $1{,}000$ steps\\

GAE parameter $\lambda_{\mathrm{gae}}$
& $0.95$
& $0.95$
& $0.95$\\

PPO epochs per update
& $4$
& $10$
& $10$\\

Minibatch size
& $1{,}024$
& $512$
& $512$\\

Number of minibatches
& $32$
& $32$
& $32$\\

Parallel environments
& $2{,}048$
& $2{,}048$
& $2{,}048$\\

Unroll length
& $5$
& $20$
& $64$\\

Entropy coefficient
& $10^{-2}$
& $10^{-3}$
& $10^{-2}$\\

Reward scaling
& $10.0$
& $10.0$
& $10.0$\\

Policy / value network
& \multicolumn{3}{c}{MLP $(64, 64)$, Swish}\\

Observation normalization
& Yes
& Yes
& Yes\\

Advantage normalization
& Yes
& Yes
& Yes\\
\bottomrule
\end{tabular}
\end{table*}

Results are reported as mean $\pm$ standard deviation across three seeds.

\begin{figure*}[h]
\centering
\includegraphics[width=\textwidth]
{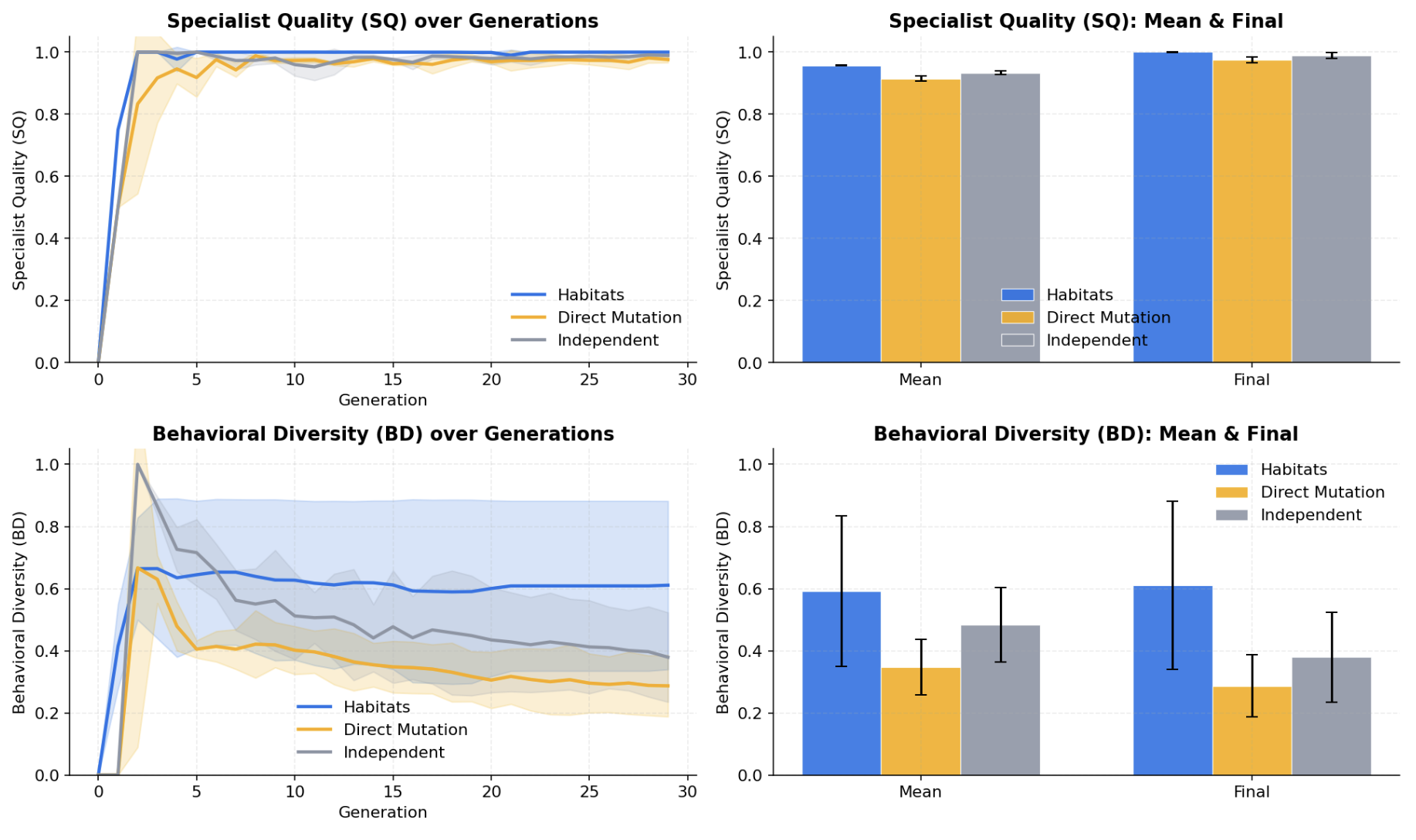}
\caption{
\textbf{Extended evolutionary horizon in Card Sandbox.}
SQ and BD across $30$ generations, together with their trajectory means and
final-generation values. Shaded regions and error bars indicate
$\pm 1$ standard deviation across three seeds.
}
\label{fig:card_extended_horizon}
\end{figure*}

Figure~\ref{fig:card_extended_horizon} shows that all three methods reach
consistently high SQ before the original $15$-generation horizon. Habitats'
BD stabilizes near $0.6$ and remains substantially higher throughout the
additional generations, finishing at $0.6109 \pm 0.2710$, compared with
$0.2870 \pm 0.0991$ for Direct Mutation and $0.3791 \pm 0.1440$ for
Independent Generation.

\begin{figure*}[t]
\centering
\includegraphics[width=\textwidth]
{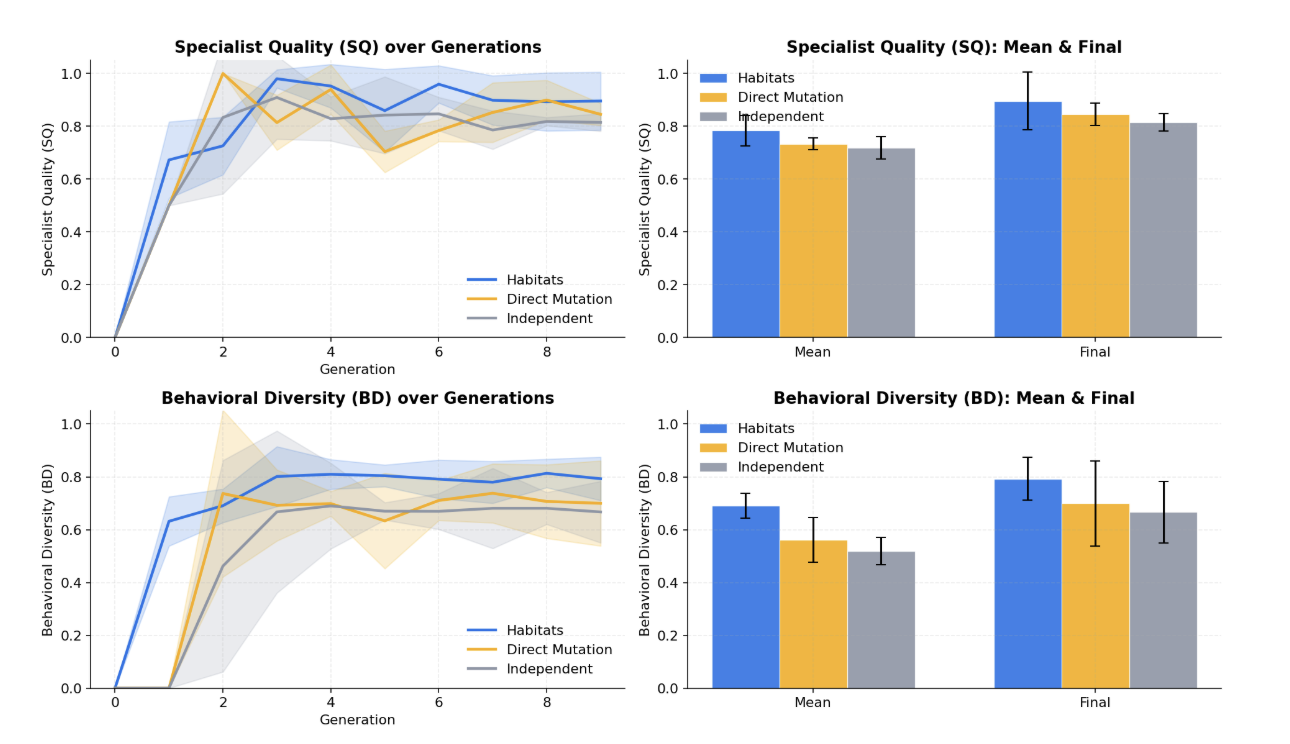}
\caption{
\textbf{Extended evolutionary horizon in Cart-Pole.}
SQ and BD across $10$ generations, with means and final values.
}
\label{fig:cartpole_extended_horizon}
\end{figure*}

\begin{figure*}[t]
\centering
\includegraphics[width=\textwidth]
{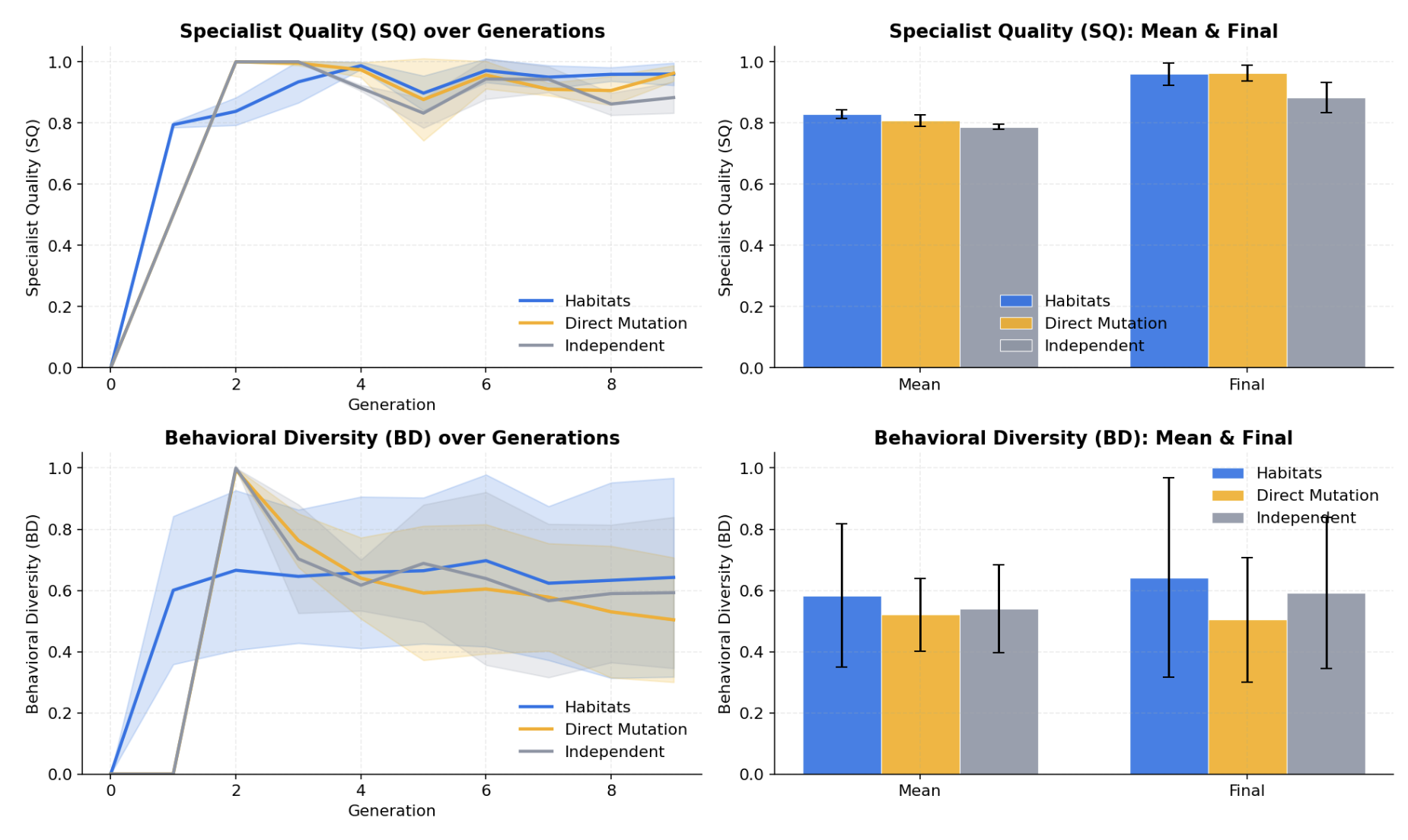}
\caption{
\textbf{Extended evolutionary horizon in Hopper.}
SQ and BD across $10$ generations, with means and final values.
}
\label{fig:hopper_extended_horizon}
\end{figure*}

\begin{figure*}[t]
\centering
\includegraphics[width=\textwidth]
{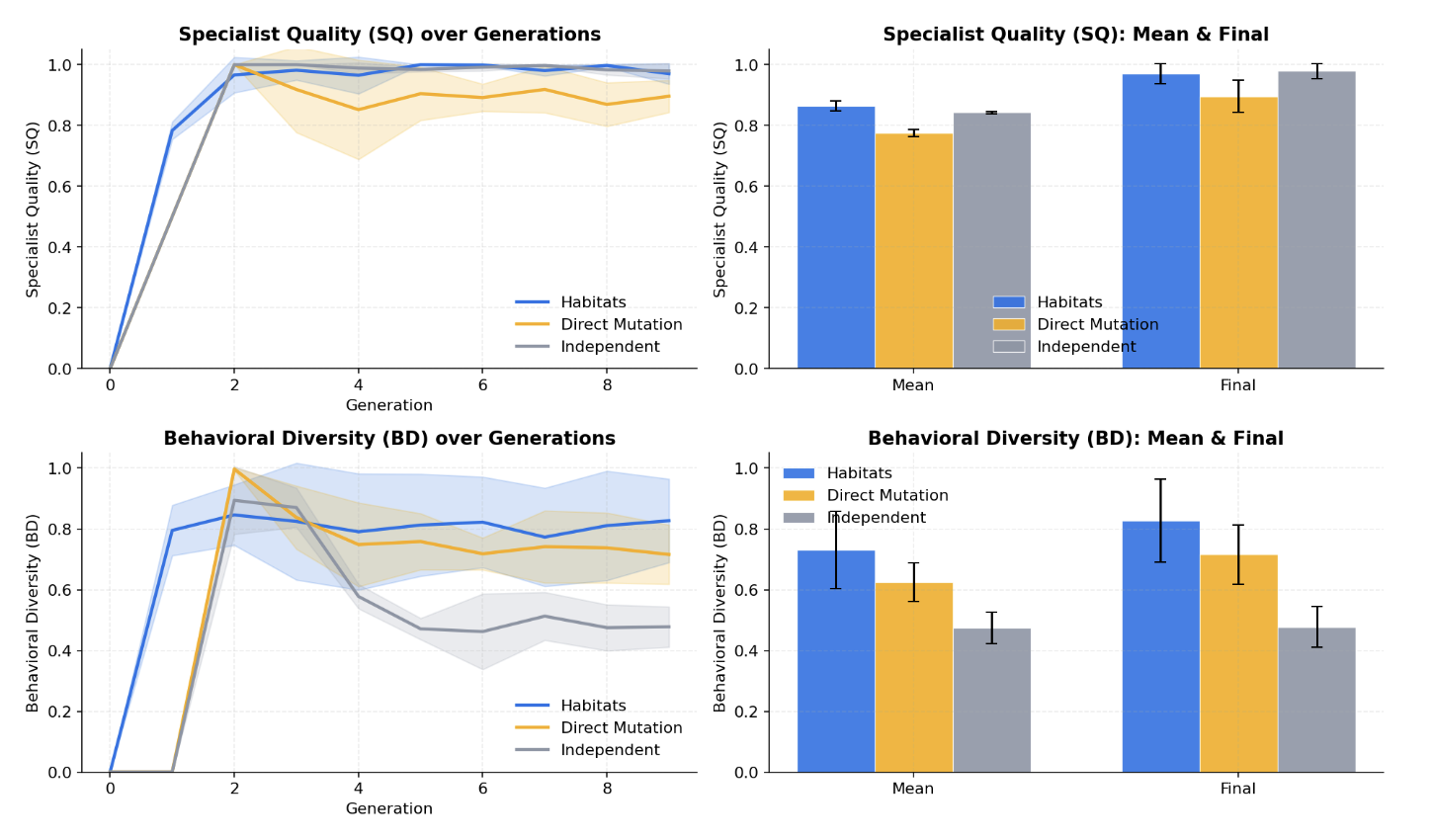}
\caption{
\textbf{Extended evolutionary horizon in Cheetah.}
SQ and BD across $10$ generations, with means and final values.
}
\label{fig:cheetah_extended_horizon}
\end{figure*}

Figures~\ref{fig:cartpole_extended_horizon}--\ref{fig:cheetah_extended_horizon}
show a consistent pattern across the continuous-control environments. SQ
enters a stable, high-performing regime by the original $6$-generation
horizon, while Habitats maintains the highest BD throughout the additional
generations. Across all four environments, extending evolution therefore
preserves the principal comparison between methods, supporting the original
horizons as sufficient for evaluation.

%%%%%%%%%%%%%%%%%%%%%%%%%%%%%%%%%%%%%%%%%%%%%%%%%%%%%%%%%%%%%%%%
% Language-Model Backbone Ablation
%%%%%%%%%%%%%%%%%%%%%%%%%%%%%%%%%%%%%%%%%%%%%%%%%%%%%%%%%%%%%%%%
\clearpage
\section{Language-Model Backbone Ablation}
\label{app:model_ablation}

We evaluate the sensitivity of Habitats to the underlying language-model
backbone in Card Sandbox. We compare \texttt{gemini-3.5-flash},
\texttt{gemini-3.8-flash}, and \texttt{gemini-4-argon}, using the same
backbone for both the Planner and Mutator. All other prompts, evolutionary
settings, policy-training budgets, and evaluation procedures are held fixed.
Each configuration is evolved for $30$ generations, and results are reported
as mean $\pm$ standard deviation across three seeds.

\begin{figure*}[h]
\centering
\includegraphics[width=\textwidth]{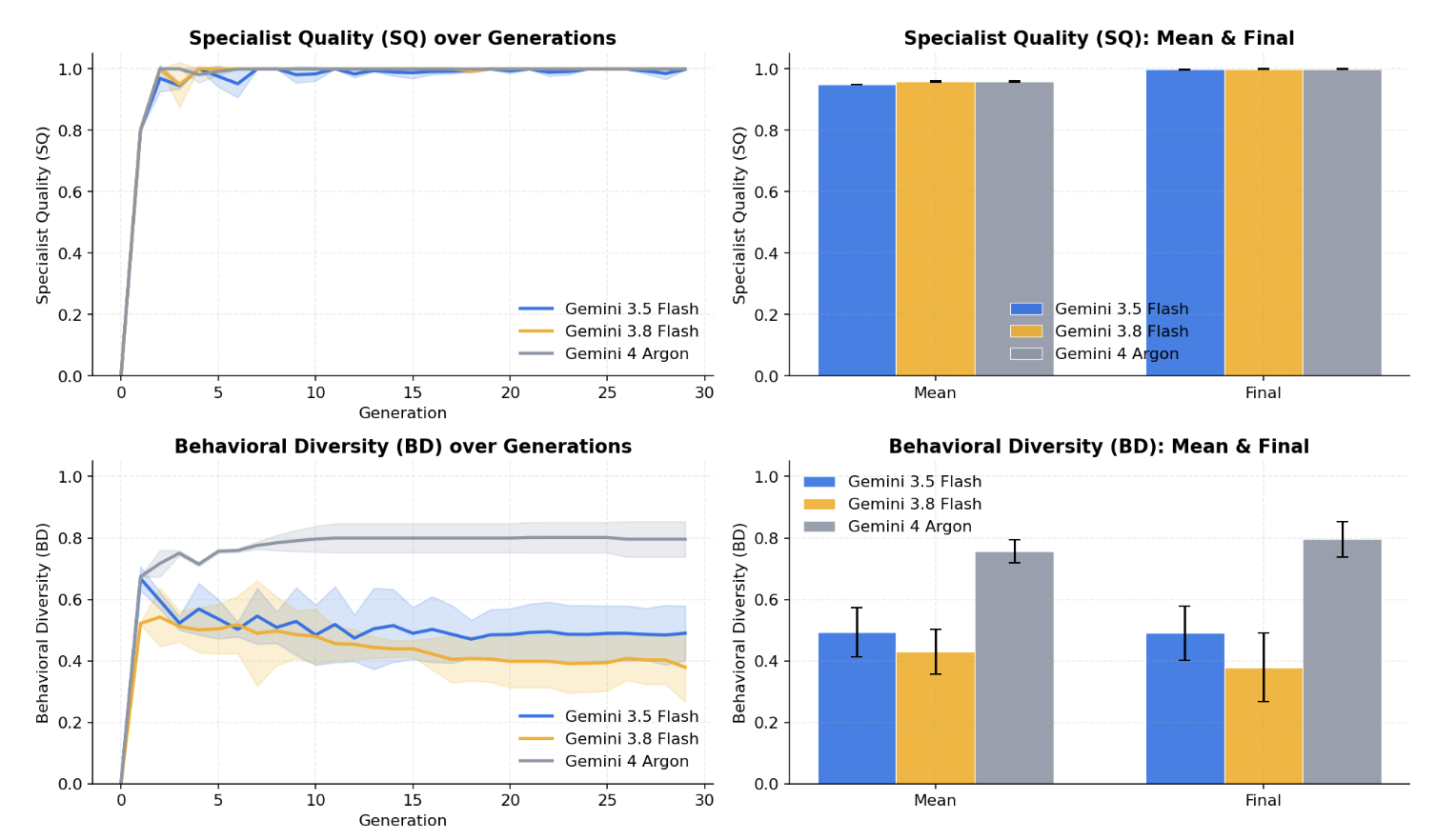}
\caption{
\textbf{Language-model backbone ablation in Card Sandbox.}
Left panels show Specialist Quality (SQ) and Behavioral Diversity (BD) across
$30$ evolutionary generations; shaded regions indicate $\pm 1$ standard
deviation across three seeds. Right panels report their means across
generations and final-generation values. All three backbones achieve
near-perfect final SQ, while Gemini 4 Argon maintains substantially greater
BD without sacrificing specialist quality.
}
\label{fig:model_backbone_ablation}
\end{figure*}

\begin{table*}[h]
\centering
\caption{
\textbf{Complete language-model backbone ablation results.}
Mean SQ and BD are averaged over the evolutionary trajectory, while final SQ
and BD are computed on the final population. Entries report mean $\pm$
standard deviation across three seeds. The highest mean in each column is
shown in bold.
}
\label{tab:model_ablation}
\small
\setlength{\tabcolsep}{5pt}
\begin{tabular}{lcccc}
\toprule
\textbf{Backbone}
& \textbf{Mean SQ}
& \textbf{Final SQ}
& \textbf{Mean BD}
& \textbf{Final BD} \\
\midrule
Gemini 3.5 Flash
& \(0.9494 \pm 0.0002\)
& \(0.9979 \pm 0.0007\)
& \(0.4929 \pm 0.0801\)
& \(0.4899 \pm 0.0888\) \\

Gemini 3.8 Flash
& \(0.9577 \pm 0.0026\)
& \(\boldsymbol{0.9996 \pm 0.0006}\)
& \(0.4294 \pm 0.0729\)
& \(0.3788 \pm 0.1127\) \\

Gemini 4 Argon
& \(\boldsymbol{0.9586 \pm 0.0020}\)
& \(0.9995 \pm 0.0007\)
& \(\boldsymbol{0.7569 \pm 0.0368}\)
& \(\boldsymbol{0.7959 \pm 0.0577}\) \\
\bottomrule
\end{tabular}
\end{table*}

Figure~\ref{fig:model_backbone_ablation} and
Table~\ref{tab:model_ablation} show that all three backbones rapidly reach
near-perfect SQ and maintain it throughout evolution. The principal difference
is behavioral diversity: Gemini 4 Argon achieves $62\%$ higher final BD than
Gemini 3.5 Flash and $110\%$ higher final BD than Gemini 3.8 Flash. As shown
by the evolutionary trajectories in Figure~\ref{fig:model_backbone_ablation},
this advantage emerges early and persists throughout evolution rather than
arising from a single final-generation measurement. These results show that
Habitats can translate stronger Planner and Mutator capabilities into a
substantially broader repertoire of distinct, learnable behaviors.

\end{document}